\documentclass{article}
\usepackage{goodfire/goodfire}

\usepackage{microtype}
\usepackage{hyperref}
\usepackage{url}
\usepackage{booktabs}
\usepackage{amsmath}
\usepackage{cleveref}
\usepackage{macros}
\usepackage{amssymb}  %
\usepackage{bm}       %

\providecommand{\topk}{k}
\providecommand{\blocksize}{b}

\usepackage{booktabs}
\usepackage{etoolbox}
\usepackage{lineno}
\usepackage[color=pink!20,textsize=scriptsize]{todonotes}

\hypersetup{colorlinks=true, citecolor=primary, linkcolor=primary, urlcolor=primary}

\providecommand{\aut}[1]{\textbf{#1}}
\providecommand{\af}[1]{{\small #1}}

\providecommand{\afn}[1]{\textcolor{antique}{$^{#1}$}}

\togglefalse{goodfireonly}  %

\newcommand{\paperauthors}{%

\authorentry{Aiden Swann}{\costar,\bullet,\extaff{a}} \authorsep
\authorentry{Mozes Jacobs}{\costar,\bullet,\extaff{b}} \authorsep
\authorentry{Mathilde Papillon}{\costar,\bullet,\extaff{c}} \\[0pt]

\authorentry{Siddharth Boppana}{\bullet} \authorsep
\authorentry{David Chanin}{\bullet} \authorsep
\authorentry{Dron Hazra}{\bullet} \authorsep
\authorentry{Usha Bhalla}{\bullet} \authorsep

\authorentry{Curt Tigges}{\bullet} \authorsep \\[0pt]
\authorentry{Mac Schwager}{\extaff{a}} \authorsep
\authorentry{Owen Lewis}{\bullet} \authorsep

\authorentry{Jack Merullo}{\bullet} \authorsep
\authorentry{Ekdeep Singh Lubana}{\bullet} \authorsep \\[0pt]
\authorentry{Atticus Geiger}{\bullet} \authorsep

\authorentry{Thomas Fel}{\colead,\bullet} \authorsep
\authorentry{Matthew Kowal}{\colead,\bullet} \\[4pt]

$^\costar$\af{Core contributors} \authorsep
\vspace{2mm}
$^\colead$\af{Equal senior contribution} \\
\goodfireaff \\
\vspace{1mm}
\repolink{https://github.com/goodfire-ai/robotics_multi_dimensional}
\\
\vspace{1mm}
\extline{%
  \afflabel{\bullet}{Goodfire}\authorsep
  \afflabel{a}{Stanford University}\authorsep
  \afflabel{b}{Harvard University}\authorsep
  \afflabel{c}{University of California, Santa Barbara}\authorsep
}
\vspace{-4mm}
}

\author{\paperauthors}

\title{Multi-Dimensional Featurizers Yield Fine-Grained Steering and Data Shaping in Robot Foundation Models}

\begin{document}
\maketitle
\vspace{-8mm}
\begin{figure}[h]
    \centering
    \vspace{8mm}
    \includegraphics[width=.82\linewidth]{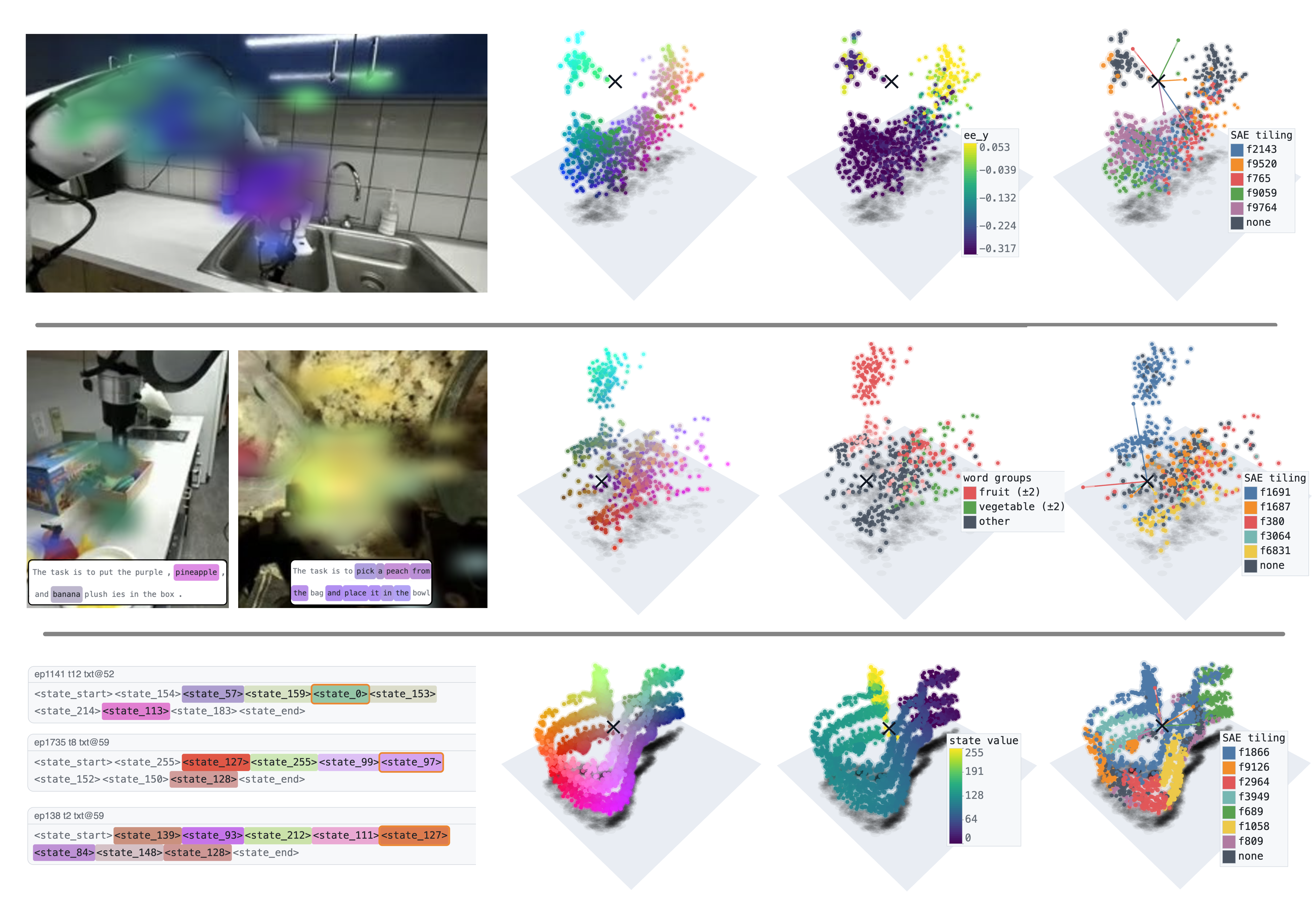}
    \vspace{-3mm}
    \caption{{\small\textbf{Block-sparse featurizers (BSFs) recover multi-dimensional robotics concepts.} Examples from vision-language-action models encode end-effector position (top), fruit-and-vegetable semantics (middle), and proprioceptive state (bottom). Each point is a token's 3D feature code, colored by PCA coordinates, task or semantic labels, and sparse-autoencoder (SAE) assignments across columns. Activations from MolmoAct2 L25, L21, L13 (top to bottom). Left overlays share the PCA colors. SAE features tile distinct regions within each block, showing how BSFs group related variation into a single feature.}}
    \label{fig:headline}
    \vspace{-3mm}
\end{figure}

\begin{abstract}

Can better feature discovery improve control over robot behavior through internal interventions and training data? Recent work in vision and language models reveals multidimensional features, motivating block-sparse featurizers (BSFs), which represent each feature in a low-dimensional subspace. We investigate whether recovering this structure provides an interpretable interface for steering vision-language-action models and shaping the data from which they learn.
We show that BSFs recover several multi-dimensional task-relevant features underlying the model's computation (e.g., perception, language, proprioceptive state, episode phase, and control variables). We causally validate these features through fine-grained steering: targeted changes in visual perception, semantic representation, and spatial localization produce predictable changes in downstream action prediction. Finally, we use feature-based data attribution to isolate where a specific behavior (failure recovery) arises in the training dataset, identifying fewer than 1.4\% of training tokens whose masking is sufficient to remove the targeted behavior. Overall, our results show how matching featurizers to representation geometry can provide more effective tools for understanding and changing robot behavior.

\end{abstract}

\section{Introduction}
\label{sec:intro}
\vspace{-0.2cm}
A vision-language-action (VLA) model maps visual observations, language instructions, and robot state to actions in the physical world \citep{intelligence2025pi05, lee2026molmoact, fang2026molmoact2, black2024pi0, brohan2023rt, kim2024openvla, deshpande2026molmob0t}. Although today's VLAs rarely perform consequential real-world tasks, fully learned robotics models may soon see widespread deployment. Yet little is known about the computations underlying their actions. Understanding these internal representations will be essential for safe deployment.

One way to understand the internal representation of a learned model is through the perspective of \emph{features}, which are recurring internal variables that encode concepts, states, relationships and ultimately underlie the process by which the model converts input to output. A promising line of work, e.g., sparse autoencoders (SAEs), leverages the \emph{linear representation hypothesis} (LRH) \citep{park2023linear,huben2024sparse,gao2025scaling, elhage2022toy} and assume that activations are composed of a sparse set of vectors. These methods have been well studied in language and vision modalities \citep{huben2024sparse,templeton2024scaling,gao2025scaling}. Examining models in the basis of the features that they express is useful for a plethora of problems, such as to study generalization \citep{gallifant2025sparse,olson2025probing}, steer model behavior \citep{templeton2024scaling, prasad2026features, turner2023steering, rimsky2024steering}, attribute behaviors to training data \citep{kowal2026concept, bergen2026anatomy, koh2017understanding}, and identify safety-relevant computation \citep{templeton2024scaling,marks2025auditing}.

However, a featurizer is ultimately a hypothesis about the data-generating process of the activations \citep{fel2026bsf}. The LRH is particularly questionable for visual domains involving perception, since continuous factors such as viewpoint, illumination, position, and motion naturally generate low-dimensional manifolds that are reflected in learned representations \citep{cohen2020separability,kvinge2024internal,fel2026bsf, engels2025not,michaud2025understanding}. Despite this, the majority of work in VLA interpretability assumes the LRH, with varying degrees of success. Early studies showed that specific MLP directions can offer limited control of variables like speed and height \citep{haon2025mechanistic}. Complementary studies use linear probes and activation interventions to characterize the contributions of visual, language, and action pathways and identify representations of environmental dynamics and task progress \citep{molinari2025emergent,grant2026not,bhardwaj2026decoding, buurmeijer2026observing}. More recently, SAEs have been applied to recover semantic concepts, motion primitives, task structure, and episode-specific memorization, demonstrating that interventions on individual features can alter policy behavior \citep{swann2026sparse,khan2025controlling,jin2026event, grant2026not}.Prior work has improved our understanding of VLA policies, but precise behavioral steering remains challenging, with interventions often producing inconsistent, off-target, or weak effects. Much of this work relies on pooled visual activations or small activation corpora, leaving open whether training SAEs on individual token activations at greater scale improves feature discovery and steering.

In this work, we apply recent methods for multi-dimensional feature discovery---block-sparse featurizers (BSFs) \citep{fel2026bsf}---to frontier open-source VLA models, training BSFs and baseline SAEs on the largest corpora of unpooled VLA token activations to date. BSFs represent features as sparse low-dimensional blocks \citep{yuan2006model,eldar2009block,jenatton2010structured,bach2012structured} rather than single directions, making them a natural fit for robotic representations. Across models and layers, we recover a broad vocabulary of task-relevant features spanning visual perception, language, proprioceptive state, task variables and episode progress. BSFs outperform matched SAEs on recovering and intervening on task-relevant structure, suggesting that the additional geometry is useful rather than simply excess representational capacity. We then causally validate them through fine-grained steering experiments across multiple layers, inducing targeted changes at several stages of the policy: what is perceived in the observation, which object is selected, where that object is localized. Finally, we perform feature-based attribution to connect interpretable behaviors directly to the episodes in the fine-tuning dataset that shaped them. Related robot data-curation methods select demonstrations with datamodels, influence estimates, or mutual information \citep{dass2026datamil,agia2025cupid,hejna2025robot}. Using a `failure-recovery' feature as as case-study, we isolate fewer than 1.4\% of fine-tuning tokens in DROID whose influence is disproportionately associated with this behavior. 
Our contributions are:
\begin{enumerate}[leftmargin=*, itemsep=3pt, parsep=0pt, topsep=3pt]
\item \textbf{Comparing block-sparse and linear featurizers for VLAs.}
We systematically compare BSFs and SAEs in VLAs. BSFs recover features spanning perception, language, proprioception, task state and episode progress, better capturing computational geometry qualitatively and quantitatively. We also strengthen VLA SAE baselines with substantially larger, unpooled activation corpora to better evaluate multidimensional features.

\item \textbf{Causal validation through fine-grained steering.}
Across multiple layers, and computational regimes, we causally validate recovered features by steering perception, semantics, target-object selection, geometric localization, and downstream robot behavior.

\item \textbf{Feature-based dataset attribution and shaping.}
We trace behaviors captured by BSF features to the data that caused them. For failure recovery, masking $\sim1.4\%$ of fine-tuning tokens qualitatively transforms robot behavior, demonstrating precise control over robot datasets and learned policies.
\end{enumerate}

\section{From Sparse Directions to Sparse Subspaces}
\label{sec:preliminaries}

\begin{wrapfigure}{r}{0.46\textwidth}
    \centering
    \vspace{-4mm}
    \includegraphics[width=\linewidth]{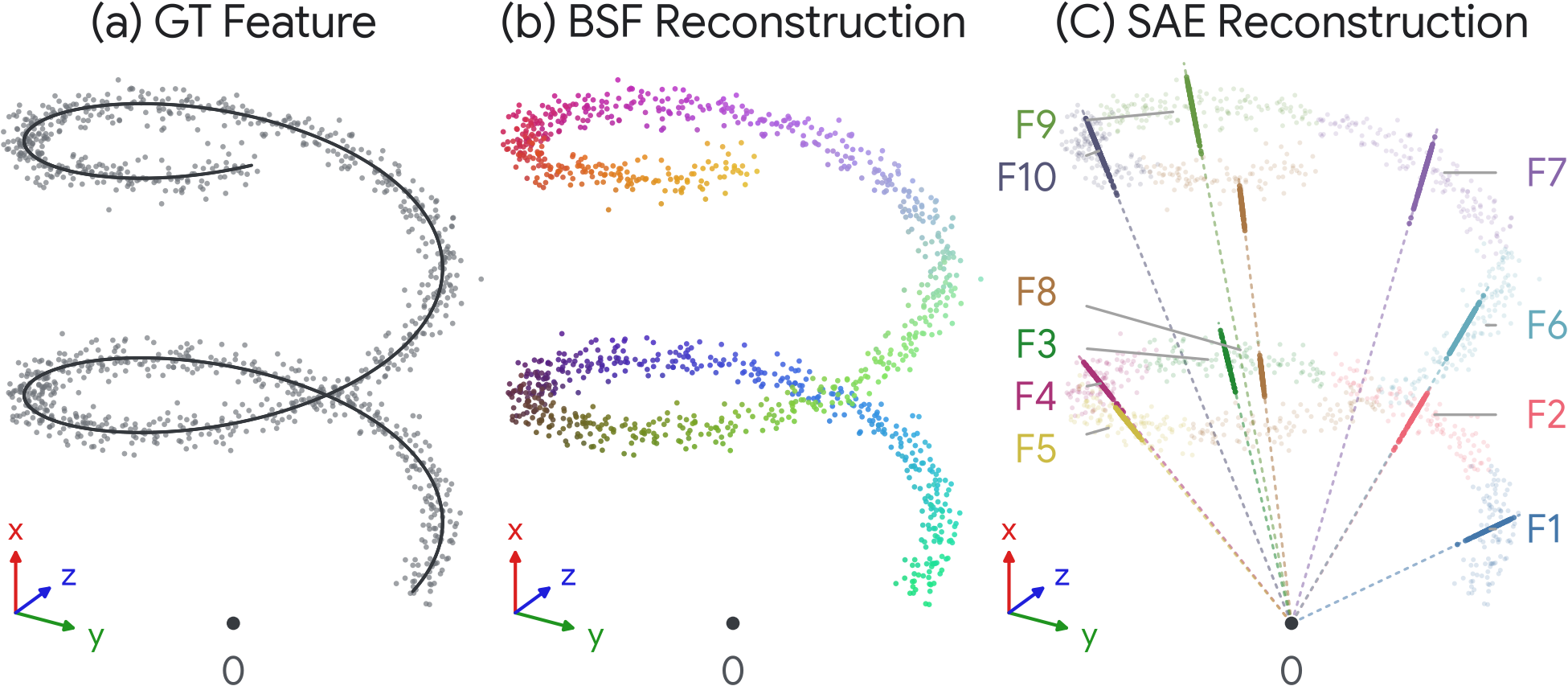}
    \caption{\textbf{BSFs better capture low-dimensional manifolds.} (b) One three-coordinate BSF block reconstructs the helix. (c) Many individual SAE features tile the helix.}
    \label{fig:bsf-helix}
    \vspace{-5mm}
\end{wrapfigure}

The goal of a featurizer is to factorize model activations into a set of interpretable concepts~\citep{huben2024sparse, bricken2023monosemanticity, gao2025scaling, bussmann2024batchtopk, rajamanoharan2024jumping, fel2025archetypal, hindupur2026projecting, bussmann2025learning}. Moreover, each featurizer also implicitly encodes a hypothesis about how activations are generated~\cite{hindupur2026projecting}. Choosing a featurizer for a VLA therefore requires matching this geometry to the variables the policy represents and computes. In this section, we first describe the feature unit assumed by sparse autoencoders, generalize it to block-sparse subspaces, explain their selection, and define the reconstruction metric shared by both featurizers.

\paragraph{Notation.}
For a positive integer $n$, we write $[n]=\{1,\ldots,n\}$. Vectors are bold lowercase and treated as row vectors, and matrices are bold uppercase. For $\bm{v}\in\mathbb{R}^{n}$, $\|\bm{v}\|_2$ is the Euclidean norm and $\|\bm{v}\|_0=|\{i\in[n]:v_i\neq 0\}|$ counts nonzero coordinates. For a matrix $\bm{A}$, $\|\bm{A}\|_{\mathrm{F}}$ is the Frobenius norm. We write $\bm{x}\in\mathbb{R}^{d}$ for an activation and $\bm{z}\in\mathbb{R}^{m}$ for its feature code. The coordinates of $\bm{z}$ are partitioned into $G$ disjoint groups $\bm{\mathcal{I}}=\{I_1,\ldots,I_G\}$ of size $\blocksize$, so that $m=G\blocksize$. The subvector $\bm{z}_g := \bm{z}_{I_g}
\in \mathbb{R}^{\blocksize}$ contains the latent coordinates associated with feature $g$. For a decoder $\bm{D}\in\mathbb{R}^{m\times d}$, the corresponding rows form the decoder block $\bm{D}_g:=\bm{D}_{I_g,:}\in\mathbb{R}^{\blocksize\times d}$. We omit decoder biases. The two featurizers below differ only in the value of $\blocksize$, and we start with the smallest one.

\paragraph{Sparse autoencoders.}\label{subsec:sae_prelim}
Sparse autoencoders use $\blocksize=1$ and $m=G$. Each feature has one coordinate $z_g$ and one decoder direction $\bm{d}_g\in\mathbb{R}^{d}$, and the reconstruction is
\begin{equation}
    \hat{\bm{x}}
    = \sum_{g=1}^{G} z_g(\bm{x})\,\bm{d}_g
    = \bm{z}(\bm{x})\bm{D},
    \qquad
    \|\bm{z}(\bm{x})\|_0 \leq \topk .
\end{equation}

The decoder range is a union of spans $\operatorname{span}\{\bm{d}_g : g\in S\}$ over supports $S\subseteq[G]$ of size $\topk$. Each feature therefore contributes along a fixed direction, with its coefficient determining magnitude and, for signed codes, its orientation. Several scalar features together can represent multidimensional variation, but the SAE architecture cannot group them as a single feature. The helix in Figure~\ref{fig:bsf-helix}, which is a one-dimensional curve with a three-dimensional linear span, illustrates this: an SAE tiles it with scalar features covering individual sub-arcs~\citep{bhalla2026sparse}, splitting one variable across features whose relationships must be recovered after training. This leads to a natural question: can the unit of a feature be enlarged to hold such a curve, while keeping a linear decoder?

\paragraph{Block-sparse featurizers.}\label{subsec:bsf_prelim}
The simplest answer is to allow $\blocksize>1$. A block-sparse featurizer (BSF)~\citep{fel2026bsf} replaces the scalar $z_g$ with a vector $\bm{z}_g\in\mathbb{R}^{\blocksize}$ and the direction $\bm{d}_g$ with a matrix $\bm{D}_g$. The reconstruction stays linear:
\begin{equation}
\label{eq:linearized_dgp}
    \hat{\bm{x}}
    = \sum_{g=1}^{G}\bm{z}_g(\bm{x})\bm{D}_g
    = \bm{z}(\bm{x})\bm{D}
\end{equation}
and feature $g$ now contributes within a learned subspace
\begin{equation}
    \mathcal{S}_g
    := \left\{\bm{u}\bm{D}_g:\bm{u}\in\mathbb{R}^{\blocksize}\right\},
    \qquad
    \dim(\mathcal{S}_g)=\operatorname{rank}(\bm{D}_g)\leq\blocksize .
\end{equation}

An activation with $\topk$ active features lies in the Minkowski sum $\mathcal{S}_{g_1}+\cdots+\mathcal{S}_{g_{\topk}}$~\citep{fel2026into}. The decoder range becomes a union of sums of subspaces instead of a union of sums of rays. The observed contributions need not fill $\mathcal{S}_g$. They can occupy a curve, a surface, or another structured subset $\mathcal{M}_g\subseteq\mathcal{S}_g$, allowing one block to hold the helix, or a multi-dimensional variable such as gripper aperture or end-effector position. Importantly, the block dimension does not fix the intrinsic dimension of $\mathcal{M}_g$ (analyzed in Section~\ref{sec:results}) and only \textit{bounds} the dimension above. 

A larger unit also changes what can be read from an active feature. Unlike an SAE feature which provides a single scalar value, a block conveys both how strongly the feature is present via the block norm $||\bm{z}_g(\bm{x})||_2$, and where the activation lies on $\mathcal{M}_g$, i.e., the position of $\bm{z}_g(\bm{x})$ within the block, without privileging any particular basis for the block coordinates.
Specifically, for any orthogonal $\bm{Q}\in O(\blocksize)$, replacing $\bm{z}_g$ with $\bm{z}_g\bm{Q}$ and $\bm{D}_g$ with $\bm{Q}^{\top}\bm{D}_g$, combined with a matching rotation of the block encoder, preserves the feature subspace, the reconstruction, and all block norms.
Unless otherwise specified, we use this freedom to display block codes in a principal-component basis in Section~\ref{sec:phenomenology}.

\paragraph{Selecting features jointly.}
BSFs assume that a block is active or inactive as a whole. If its coordinates could be selected independently, the block norm would no longer measure the presence of one feature, and the block would decompose back into scalar features. Sparsity must therefore act on groups rather than on coordinates, so that the coordinates of a selected block vary together and unselected blocks are exactly zero. We count active features with the group pseudo-norm
\begin{equation}
    \|\bm{z}\|_{0,\bm{\mathcal{I}}}
    := \sum_{g=1}^{G}\mathbf{1}\!\left(\|\bm{z}_g\|_2>0\right),
\end{equation}
which reduces to $\|\bm{z}\|_0$ when $\blocksize=1$. We use the Vanilla BSF of~\citet{fel2026bsf}, with independently learned encoder and decoder. Its per-activation encoder is
\begin{equation}
\label{eq:vanilla_bsf}
    \bm{z}(\bm{x})
    = \Pi_{\topk}\!\left(\bm{x}\bm{W}+\bm{b}_{\mathrm{enc}}\right),
    \qquad
    \bm{W}\in\mathbb{R}^{d\times m},
    \quad
    \bm{b}_{\mathrm{enc}}\in\mathbb{R}^{m}.
\end{equation}
Here $\Pi_{\topk}$ is the Euclidean projection onto $\{\bm{z}:\|\bm{z}\|_{0,\bm{\mathcal{I}}}\leq\topk\}$. It keeps the $\topk$ blocks with largest $\ell_2$ norm and sets the others to zero. Retained coordinates are signed, and no coordinate-wise nonlinearity is applied. For $\blocksize=1$, the block norm becomes an absolute value and the featurizer becomes a TopK SAE~\citep{gao2025scaling,zhu2026abstopk}. 
All SAE baselines in this paper are this special case, trained on the same data with the same pipeline, so the block dimension is the only difference between the two featurizers. 
In practice, we train with the BatchTopK variant of $\Pi_{\topk}$~\citep{bussmann2024batchtopk}, which keeps the $B\topk$ highest-scoring blocks across a batch of $B$ activations, with each block scored by $\|\bm{a}_{ng}\|_2\|\bm{D}_g\|_{\mathrm{F}}$ to account for decoder scale. 
Because a block uses $\blocksize$ coordinates where a scalar feature uses one, we always report which budget is matched, either the number of active features $\topk$ or the number of active coordinates $\topk\blocksize$; details are given in Appendix~\ref{app:collection_training}. With the unit of a feature and its selection rule fixed, the remaining choice is the objective used to train it.

\paragraph{Reconstruction and functional reconstruction.}
We first recall briefly the usual featurizer reconstruction loss and then introduce a functional invariant that we investigate in parallel.
The standard objective is the expected squared reconstruction error,
\begin{equation}
\label{eq:rec_loss}
    \mathcal{L}_{\mathrm{rec}}
    = \mathbb{E}_{\bm{x}}\!\left(\|\bm{x}-\hat{\bm{x}}\|_2^2\right).
\end{equation}
This loss treats all directions of activation space equally: an error of a given norm costs the same wherever it points. 
However, some directions of the activation strongly affect the predicted action, while others have almost no effect on it. 
We therefore also measure reconstruction error by its effect on the policy's \textit{output}. 
Let $\bm{g}\in\mathbb{R}^{d}$ be the gradient, with respect to the activation, of the teacher-forced log-likelihood of the on policy action tokens. For an error $\bm{e}=\bm{x}-\hat{\bm{x}}$, the change in log-likelihood is, to first order, $\bm{e}\bm{g}^{\top}$. 
Averaging the square over a corpus of $N$ gradients gives
\begin{equation}
\label{eq:fisher_quad}
    \frac{1}{N}\sum_{n=1}^{N}\left(\bm{e}\bm{g}_n^{\top}\right)^2
    = \bm{e}\,\hat{\bm{F}}\,\bm{e}^{\top},
    \qquad
    \hat{\bm{F}} = \frac{1}{N}\sum_{n=1}^{N}\bm{g}_n^{\top}\bm{g}_n,
\end{equation}
where $ \hat{\bm{F}} \in \mathbb{R}^{d \times d}$ is the empirical Fisher matrix. An error along a direction of large curvature changes the predicted action and is thus heavily penalized. 
Put differently, if there exist a null space in the activations (e.g. a neuron with no effect) then the reconstruction error in the null space of $\hat{\bm{F}}$ satisfies $\bm{e}\bm{g}_n^{\top}=0$ for every $n$: it has no first-order effect on the action and does not need to be reconstructed.

\paragraph{Activation corpus and SAE baselines.}

Since our claims about block structure’s benefits are relative to an SAE, we aim for the strongest possible baseline. Prior SAEs trained on VLA activations have two limitations: the main experiments of~\citet{swann2026sparse} mean-pool activations, combining all modalities into one vector; \citet{grant2026not} and~\citet{jin2026event} avoid pooling but use small corpora—$500$k action-token activations and a few hundred simulated rollouts per SAE, respectively. Instead, we retain each image, text, and state token as a separate activation, yielding approximately $52$M training activations collected uniformly across 45,344 episodes, roughly two orders of magnitude more than the pooled activations in the next largest dataset from~\citet{swann2026sparse}. All our featurizers train on this corpus, so differences in results can be attributed purely to featurizer structure.

\section{Results}
\label{sec:results}
Given the featurizers covered in \Cref{sec:preliminaries}, we now turn to the features they discover and the understanding and control they admit. We first perform a quantitative comparison of BSFs to SAEs (\Cref{sec:sae_compare}) and causally validate the discovered features through open- and closed-loop steering experiments (\Cref{sec:validation_steering}).
We then review a broad range of feature phenomenology, spanning visual, language, control, and proprioceptive features (\Cref{sec:phenomenology}). 
Finally, we demonstrate an application of BSF features for the goal of behavior modification through data attribution (\Cref{sec:validaton_data}).

\subsection{Quantitative Comparison to SAEs}
\label{sec:sae_compare}

\begin{figure}[t]
    \centering
    \begin{minipage}[c]{0.5\linewidth}
        \includegraphics[width=\linewidth]{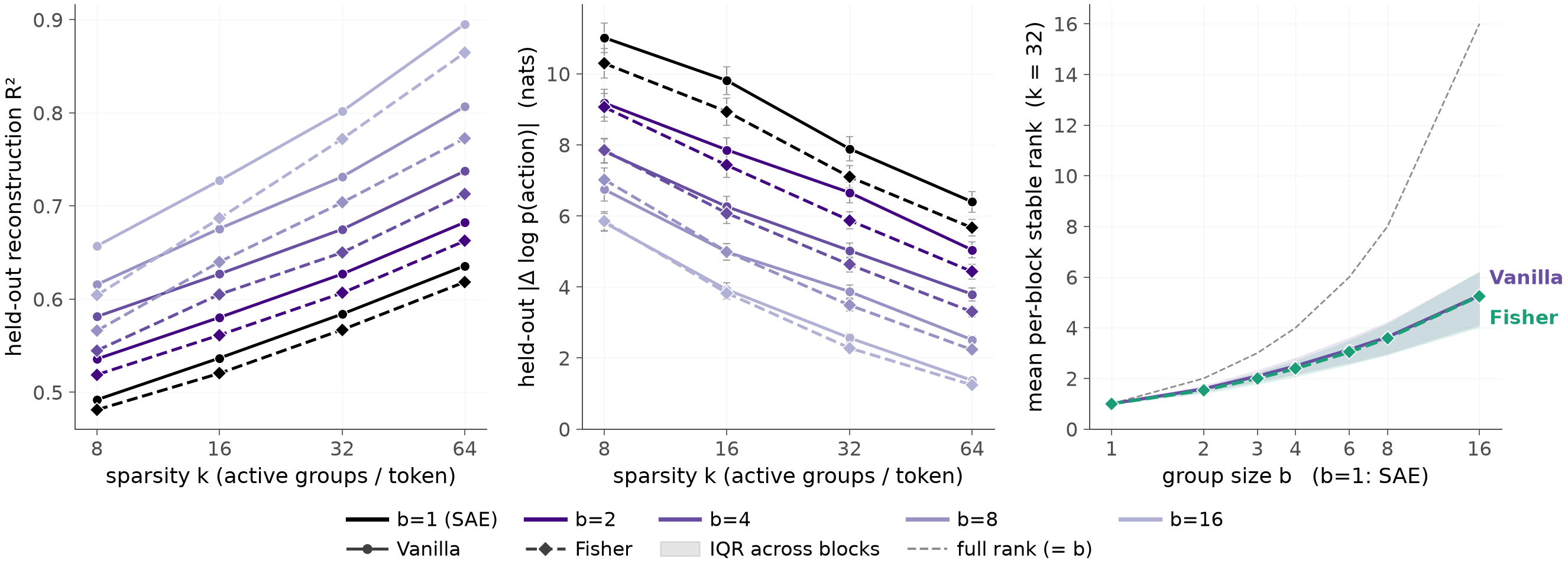}
    \end{minipage}\hfill
    \begin{minipage}[c]{0.49\linewidth}
        \includegraphics[width=\linewidth]{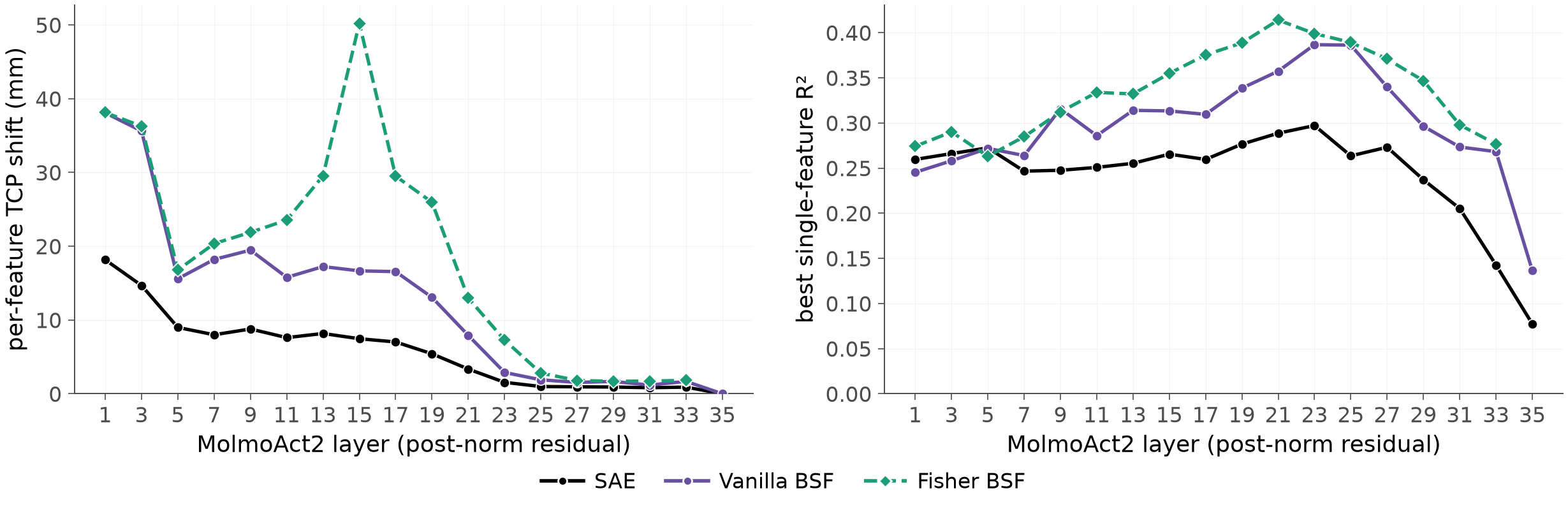}
    \end{minipage}
    \caption{\textbf{Reconstruction, dimensionality, and behavioral relevance of SAE and BSF features in MolmoAct2.}
    \emph{Left three panels (layer 13):} Held-out reconstruction $R^2$ and absolute recorded-action log-likelihood change versus active blocks $\topk$ (lower change is better; mean $\pm$ SEM, 48 episodes), and within-block stable rank versus block size at $\topk=32$ (bands: interquartile range; gray: full dimension). Colors denote block size ($\blocksize=1$: SAE); solid/dashed lines denote Vanilla/Fisher training.
    \emph{Right two panels (across layers):} Three-dimensional Vanilla and Fisher BSFs versus SAEs at approximately matched latent widths and active-coordinate budgets (21 blocks versus 63 latents). Ablation effects measure mean predicted end-effector displacement over the top seven BSF blocks or 21 SAE latents by activation mass. Decodability is the best single-feature ridge-probe $R^2$ on held-out episodes, averaged over ten robot-state and task-progress variables.}
    \label{fig:quant_combined}
\end{figure}

We hypothesize that BSFs---due to their inductive prior of low-dimensional features that are smooth over time---are a more natural fit to VLAs than traditional SAEs. We perform a quantitative comparison between SAEs, Vanilla BSFs, and Fisher-weighted BSFs on MolmoAct2, and use the following three metrics: (i) \textit{activation reconstruction}, (ii) \textit{preservation of output behavior} (i.e., action likelihood), and (iii) robotics-grounded measures of \textit{causal influence} and \textit{task-variable decodability}.

Figure~\ref{fig:quant_combined} (column 1 and 2) shows held-out activation reconstruction and action fidelity (i.e., how much the teacher-forced log-likelihood of recorded actions changes when prompt activations are replaced by their reconstructions) for various methods and block sizes (note, SAEs are just a special case of BSFs where block size is one). As expected, larger blocks improve both metrics at fixed block sparsity while Fisher weighting trades lower $R^2$ for better preservation of action likelihoods.
We further estimate the effective dimensionality for the features recovered using the \textit{stable rank} of each  (\Cref{fig:quant_combined} column 3). Vanilla and Fisher BSFs both grow slower than the block dimension, reaching a mean stable rank of approximately $5.3$ for 16-dimensional blocks.

\Cref{fig:quant_combined} ((column 4 and 5) shows the comparison for the two robotics-specific metrics. First, \emph{task-variable decodability} measures how well a linear probe predicts robot state and task progress from individual features, i.e., the best held-out single-feature $R^2$ for each target, averaged across targets. Specifically, the targets are comprised of: end-effector position and displacement, episode progress, gripper state, distance to the first gripper-closure position, and whether that closure has occurred. Second, \emph{per-feature causal effect} measures the change in predicted end-effector position after ablating an individual feature, where more causal features are assumed to induce a larger change. Indeed, we observe that BSFs produce both larger per-feature ablation effects than SAEs across layers as well as higher task-variable decodability, particularly through the middle and later layers. Moreover Fisher-weighted BSFs match or exceed vanilla BSFs across both causal effect and task-variable decodability.

\subsection{Causal Validation of Features via Steering}
\label{sec:validation_steering}

\paragraph{Open-Loop Steering Results.}

\begin{figure}[t]
\centering
\begin{minipage}[c]{0.32\linewidth}
    \centering
    \includegraphics[width=\linewidth]{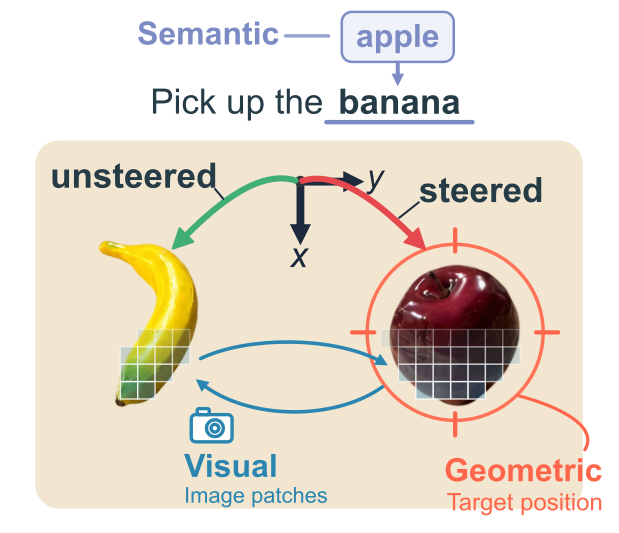}
\end{minipage}
\hfill
\begin{minipage}[c]{0.66\linewidth}
    \centering
    \small
    \setlength{\tabcolsep}{3pt}
    \renewcommand{\arraystretch}{1.08}
    \resizebox{\linewidth}{!}{%
    \begin{tabular}{@{}llrr@{}}
    \toprule
    Intervention & Featurizer &
    \shortstack{\# Features$\downarrow$} &
    \shortstack{Held-out recovery $\uparrow$} \\
    \midrule
    \textcolor[HTML]{7C8AC3}{Semantic L3}          & BSF                & 2.15  & 83.8\% \\
    Noun token           & Fisher BSF$^{*}$   & 2.56  & 78.9\% \\
                         & SAE                & 3.50  & 81.9\% \\
    \midrule
    \textcolor[HTML]{2A86B0}{Visual L7}        & BSF        & 20.40 & 84.6\% \\
    63 exo-image tokens  & Fisher BSF & 16.33 & 88.3\% \\
                         & SAE        & 36.25 & 97.3\% \\
    \midrule
    \textcolor[HTML]{FF654A}{Geometric L23}        & BSF        & 1.00  & 99.7\% \\
    Action-readout token & Fisher BSF & 1.08  & 83.9\% \\
                         & SAE        & 1.00  & 98.2\% \\
    \bottomrule
    \end{tabular}%
    }
\end{minipage}
\caption{\textbf{Fine-grained steering through three representations.}
\textcolor[HTML]{7C8AC3}{Semantic},
\textcolor[HTML]{2A86B0}{visual}, and
\textcolor[HTML]{FF654A}{geometric} interventions redirect predicted
actions from the prompted object to the other object while leaving
input images and instructions unchanged.
Results cover 52 held-out scene--instruction comparisons with six
noise seeds each.
\emph{\# Features}: selected-set size, counting three-dimensional
BSF blocks or scalar SAE features.
\emph{Held-out recovery}: fraction of the activation-patching effect
reproduced by the selected features, averaged over comparisons for
which feature discovery finds a steering set.
$^{*}$A steering set is found for only 50/52 comparisons; all other rows
cover all 52. Details in Appendix~\ref{app:steering_protocol}.}\vspace{-0.3cm}
\label{fig:steering_combined}
\end{figure}

A useful featurizer should enable targeted changes to policy behavior through a small number of feature edits that generalize to new scenes. We evaluate this with a redirection task in which two objects sit at opposite ends of a table, making the first predicted action chunk indicative of the robot's intended target. We instruct the policy to pick up one object and test the minimum number of feature edits required to redirect its reach toward the other. To scale this evaluation across hundreds of scene--method combinations, we use an automated search to identify compact feature sets that achieve redirection (see Appendix~\ref{app:steering_protocol} for full details) on a `train' set of episodes, and then hold the selected sets fixed and evaluate their effectiveness on held-out episodes.

We apply this procedure at three sites associated with distinct aspects of target grasping (Figure~\ref{fig:steering_combined}). Editing the early-layer noun token changes \emph{what} the robot is looking for; editing mid-layer image tokens changes \emph{where} it perceives each object; and editing the late-layer action-readout token changes the represented target \emph{position}. At each steering location, replacing the activations with those from the condition targeting the other object flipped the robot's target action towards the corresponding object. We use this activation-patching effect as a reference, measuring held-out recovery as the fraction of our steering relative to it. Held-out evaluation tests whether the selected feature sets generalize beyond the episodes we found them on, while task completion is evaluated separately in closed loop (Table~\ref{tab:steering_rollouts}). Full details are provided in Appendix~\ref{app:steering_protocol}.

Figure~\ref{fig:steering_combined} reports two metrics averaged over 52 scenes: steerable feature search success, the number of selected features, and held-out recovery. Each feature is one SAE latent or BSF block, and 100\% recovery means matching the trajectory shift produced by activation patching (Appendix~\ref{app:steering_protocol}). All configurations yield successful interventions on 52/52 scenes except semantic steering with Fisher-weighted BSFs, which succeeds on 50/52. Geometric steering is the most compact: a single BSF block achieves 99.7\% recovery, comparable to a single SAE latent. For semantic steering, standard BSFs require fewer features while slightly improving held-out recovery. Perceptual steering reveals a trade-off: the BSF variants require roughly 16--20 blocks, compared with 36 SAE latents, but the SAE transfers more effectively to held-out scenes. Overall, BSFs thus provide more compact semantic and perceptual interventions, although fewer feature edits do not consistently yield better transfer.

\begin{table}[t]
\centering
\small
\setlength{\tabcolsep}{3pt}
\begin{tabular}{@{}l cccc cccc@{}}
\toprule
& \multicolumn{4}{c}{SAE} & \multicolumn{4}{c}{BSF} \\
\cmidrule(lr){2-5}\cmidrule(lr){6-9}
Object pair & \# Feat. & Select & Grasp & Task & \# Feat. & Select & Grasp & Task \\
\midrule
Apple / banana & 2.0 & 90\%  & 85\%  & 75\%  & 2.5 & 100\% & 95\%   & 90\%   \\
Bowl / cable   & 2.5 & 100\% & 95\%  & 85\%  & 2.0 & 100\% & 100\%  & 100\%  \\
Mouse / towel  & 3.0 & 75\%  & 70\%  & 65\%  & 2.5 & 100\% & 95\%   & 90\%   \\
\midrule
Total          & 2.5 & 88.3\% & 83.3\% & 75.0\% & 2.3 & 100\% & 96.7\% & 93.3\% \\
\bottomrule
\end{tabular}
\caption{\textbf{Closed-loop semantic target redirection.} The policy is instructed to pick up one
object while a fixed semantic feature intervention redirects it to the other. Entries are success
rates over 20 trials per object pair (10 per direction; 60 in total): Select, the policy goes for the
redirected object; Grasp, it grasps that object; Task, it places that object in the basket.
\# Feat.: features per direction, averaged. Per-direction counts are in
Table~\ref{tab:steering_rollouts_full}.}
\label{tab:steering_rollouts}
\end{table}

\paragraph{Closed-Loop Target-Object Redirection.}
We next test whether semantic steering, whose intervention does not depend on the specific observation, persists through closed-loop execution. The robot is asked to ``pick up the [noun] and put it in the basket'' for one object, while a fixed noun-token feature intervention, selected as in \Cref{fig:steering_combined}, redirects it to the other object for the whole episode. We use three object pairs, each steered in both directions for 10 trials, and randomize the object and basket positions in every trial. BSF features redirect the policy more reliably with fewer features (\Cref{tab:steering_rollouts}): they select the redirected object in every trial and complete the task in 93.3\% of trials using 2.3 features on average, compared with 88.3\% selection and 75.0\% completion for 2.5 SAE features.

\begin{figure}[h]
    \centering
    \includegraphics[width=1.0\linewidth]{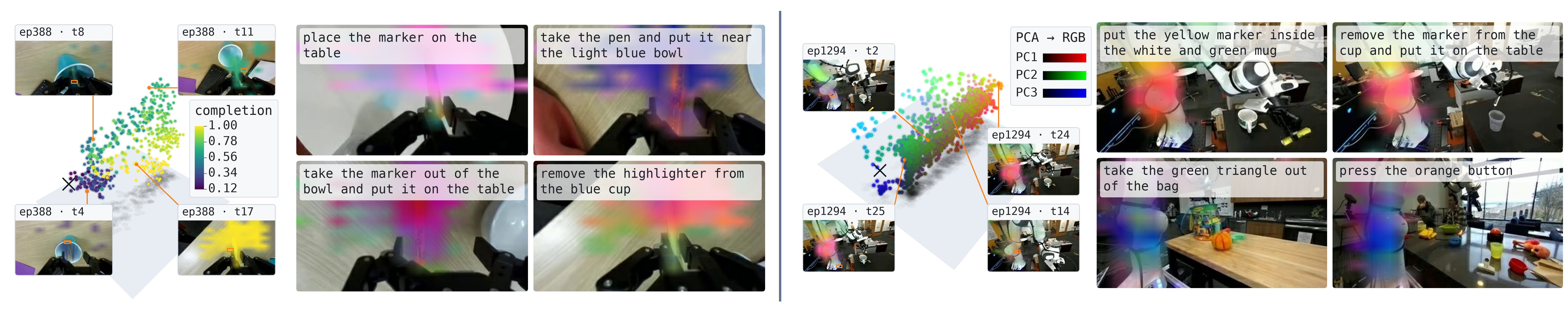}
    \caption{\textbf{Visual BSF features capture variation across task stages and robot parts.} (left) Marker feature \texttt{F2973}, with points colored by task completion. Linked images illustrate changes in appearance across a single episode; the larger overlays show the four highest-activating timesteps in the dataset. (right) Robot feature \texttt{F2571}, with PCA colors and linked images showing how different regions of the block correspond to different robot parts. Both features are Fisher-weighted BSFs from MolmoACT2 layer 7.}
    \label{fig:phen-visual}
\end{figure}

\begin{figure}[h]
    \centering
    \includegraphics[width=1.0\linewidth]{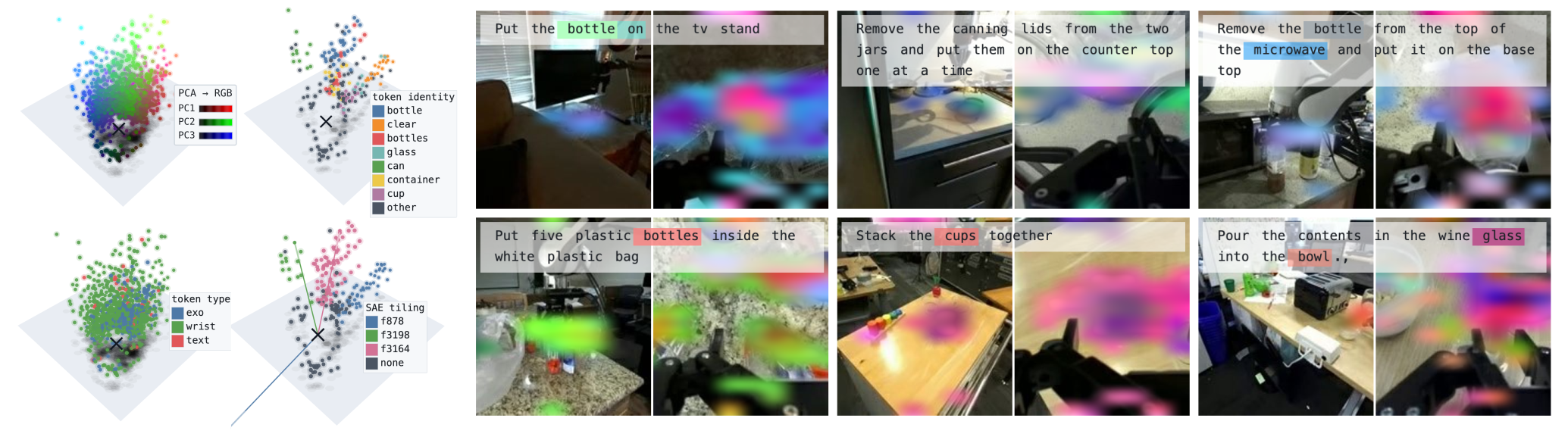}
    \vspace{-0.5cm}
    \caption{ \textbf{A single BSF feature links related container concepts across vision and language.} The left column of point clouds shows activations colored by PCA coordinates (top) and token modality (bottom). The right column shows only activating text tokens, colored by token identity (top) and SAE assignment (bottom). The feature nicely represents different types of containers across spatially distinct positions.  This feature is Vanilla BSF \texttt{F1777} from $\pi_{0.5}$ layer 3.}
    \label{fig:phen-vision-language}
\end{figure}

\subsection{Feature phenomenology}
\label{sec:phenomenology}

In this section, we highlight a representative sample of BSF features across models and layers, focusing on how variation within a block relates to interpretable properties of its inputs. Broadly, we find BSF features associated with visual, visual-language, and robot state and control concepts. Each point in the activation visualization represents the three-dimensional code of a block at an individual token where it activates and we center and rotate these codes using PCA, retaining all three dimensions.

\begin{figure}[h]
    \centering
    \includegraphics[width=0.9\linewidth]{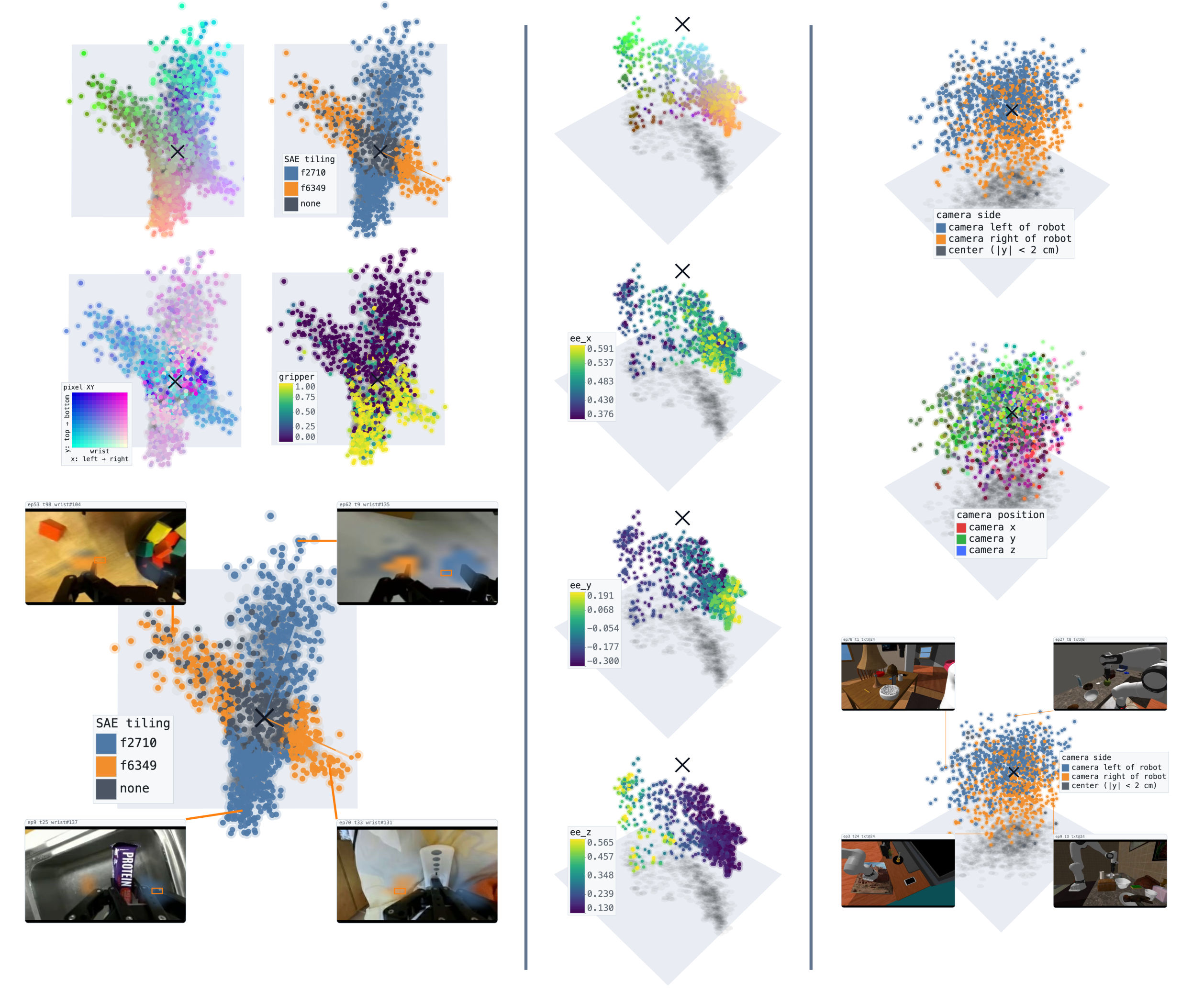}
    \vspace{-0.4cm}
    \caption{\textbf{A sample of BSF control-relevant features.} \emph{Left:} MolmoACT2 layer 13, \texttt{F937}, sampled across the dataset. Colorings show PCA coordinates, SAE assignment, image-patch position, and gripper state; linked images illustrate the gripper configurations and locations associated with selected points. \emph{Middle:} $\pi_{0.5}$ layer 7, \texttt{F2113}, sampled from a single episode and colored by PCA coordinates (top) and end-effector $x$, $y$, and $z$ position (bottom). \emph{Right:} MolmoBot layer 24, \texttt{F725}, colored by camera side and camera position in the robot frame, with four representative views.}
    \label{fig:phen-control}
\end{figure}

\paragraph{Visual Features.}
We find visual features similar to those reported by \citet{fel2026bsf} in DINOv3~\citep{simeoni2025dinov3}, now in VLA activations on robotics data (\Cref{fig:phen-visual}). These include features selective for objects in the environment and for the robot itself. In contrast to the features found in DINOv3, latent reconstruction is not part of the VLA training objective.

We show two example features in \Cref{fig:phen-visual}, one which activates on `markers' and another which activates on the robot. Both features illustrate how a block can capture concept variation within a single object category: the marker's activation geometry varies with the appearance of markers at different stages of manipulation, while the robot feature instead organizes patches from different parts of the arm. In both cases, the block captures recognizable visual content together with structure within that content. Across our models and layers we find plentiful features for a wide range of objects.

\paragraph{Visual-Language Features.}
Many blocks activate across both pixels and instruction text and are common across the tested models and layers. For example, Figure~\ref{fig:phen-vision-language} shows a container feature in $\pi_{0.5}$ that activates on external-camera, wrist-camera, and text tokens. Within the block, words referring to related containers occupy distinct regions. This organization is consistent with a shared vision--language representation that retains distinctions between members of a semantic category. The fruit-and-vegetable feature in Figure~\ref{fig:headline} provides another example of a block spanning related objects. Notably, three SAE features tile much of the displayed text-token geometry, while the BSF represents these related groups within a single block.

\paragraph{Task and Control Variables.}
Beyond visual and semantic content, we find blocks associated with robot configuration and spatial variables (Figure~\ref{fig:phen-control}). A gripper feature forms an \texttt{X}-shaped point cloud whose organization reflects both gripper opening and the side of the gripper on which the feature activates. Notably, each side is tiled by exactly one SAE feature. The BSF block brings these complementary views of a changing two-side gripper into a single feature.

A second feature varies with the end-effector's Cartesian coordinates over a single episode. This association is notable because the policy receives joint angles as proprioceptive input, rather than explicit Cartesian end-effector coordinates. The feature also primary activates on unused image token positions near the robot base. This use of image-tokens for information storage is a unique phenomenon to $\pi_{0.5}$ given its bidirectional attention mask. 

A third example comes from MolmoBot, which is trained on synthetic data. Known camera poses allow us to relate the feature's geometry to camera position relative to the robot, including whether the camera lies to its left or right. This feature captures a property of the observation viewpoint alongside the robot-state variables illustrated above.

\begin{figure}[h]
    \centering
    \includegraphics[width=0.99\linewidth]{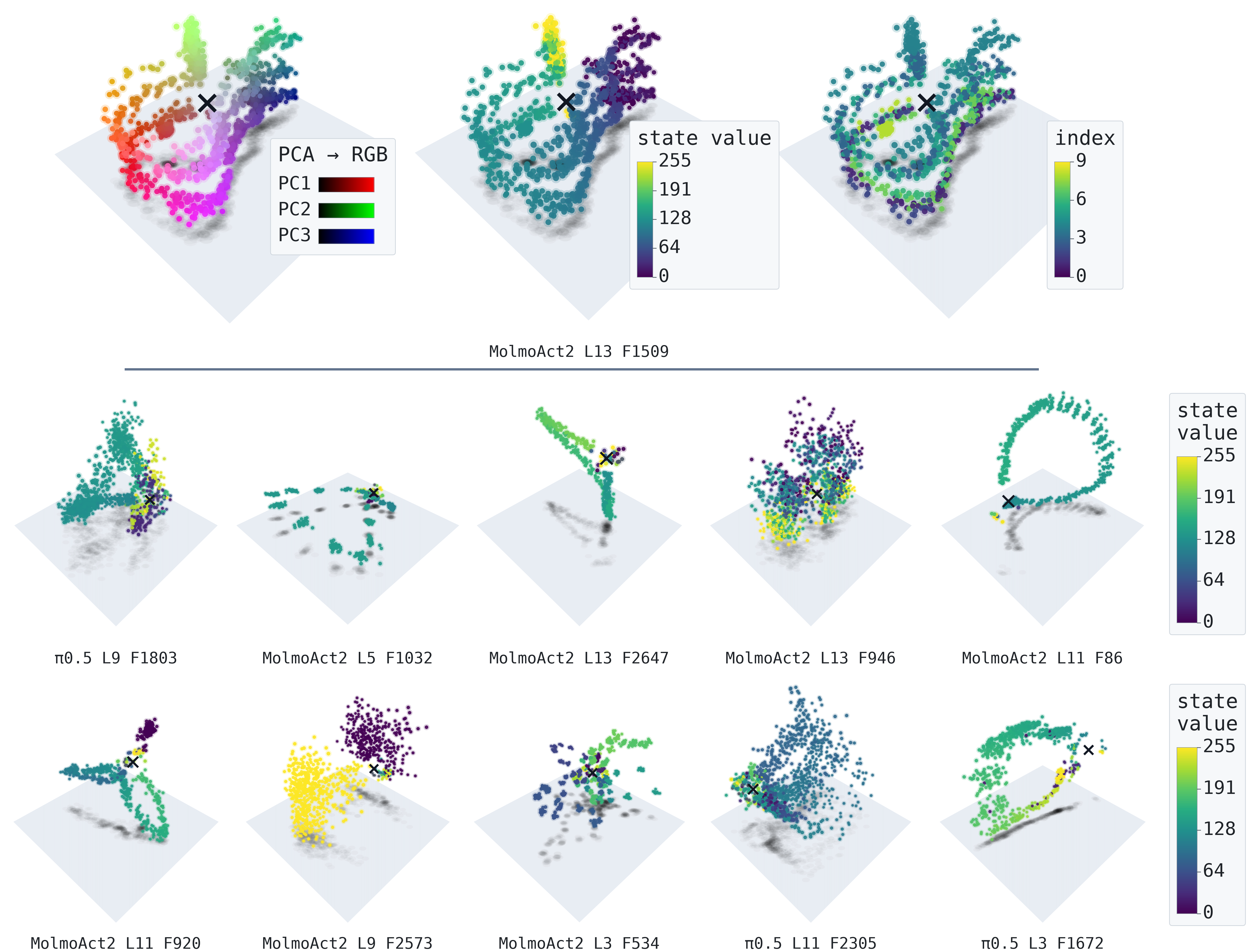}
    \caption{\textbf{Sampling of vanilla BSFs proprioception features.}
      \emph{Top:} A MolmoACT2 layer-13 feature shown with PCA coloring (left), quantized state value from 0 to 255 (middle), and state-component index (right).}
    \label{fig:phen-proprio}
\end{figure}

\paragraph{Proprioceptive Features.}
Proprioception, the robot's state information, enters the VLM backbone as part of the prompt. The two models encode these inputs differently: $\pi_{0.5}$ expresses discretized state values as numeric text, while MolmoACT2 uses dedicated discrete state tokens. Despite this difference, both models exhibit features with geometry related to proprioceptive values as shown in \Cref{fig:phen-proprio}.

The highlighted MolmoACT2 feature notably forms a complex manifold of three looping branches that converge into a common extension. Coloring by quantized state value reveals an ordering along the curves, while coloring by state-component index reveals additional structure associated with which component is represented. The block thus organizes both numeric value and component identity. Other examples exhibit discrete clusters, curved trajectories, and more diffuse clouds, demonstrating the range of geometries associated with proprioceptive inputs across layers and models.

\section{Application: Behavior Shaping via Data Attribution}
\label{sec:validaton_data}

\begin{figure}[b]
    \centering
    \includegraphics[width=\linewidth]{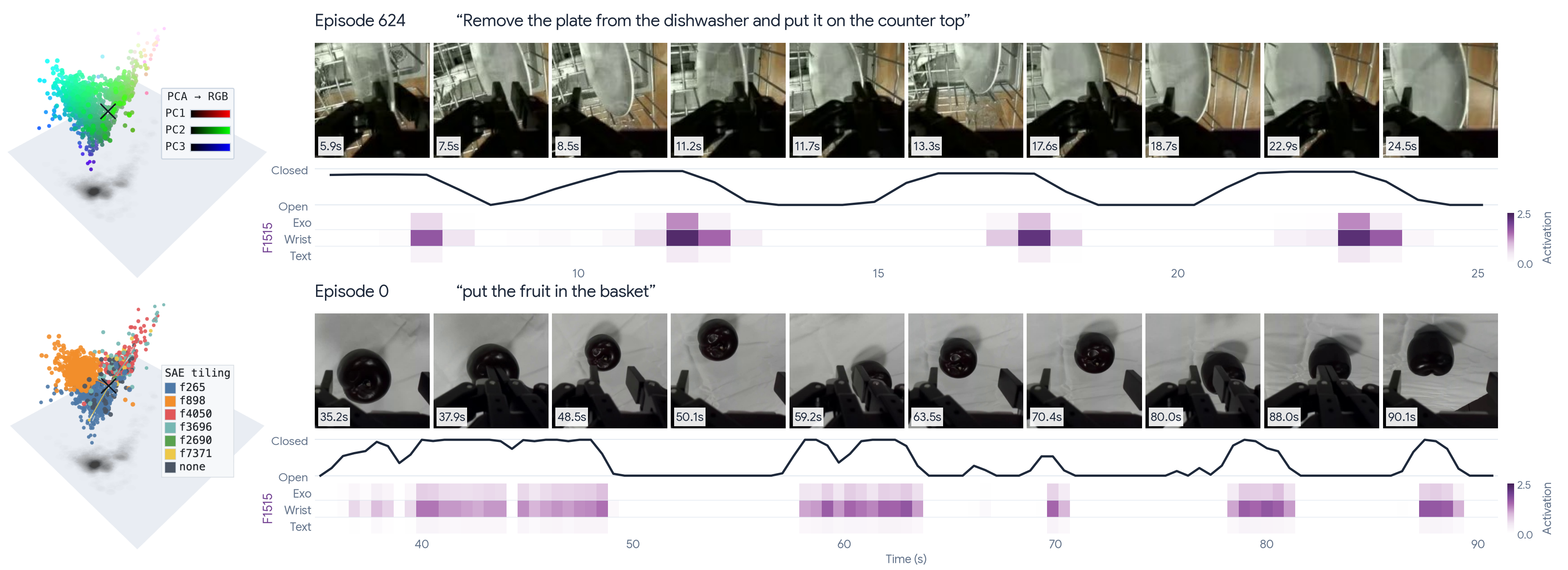}
    \vspace{-0.5cm}
    \caption{\textbf{Feature associated with failure recovery.} \emph{Left:} \texttt{F1515} block codes colored by PCA coordinates and SAE assignment. \emph{Right:} Wrist observations, gripper state, and mean feature norm across exterior-image, wrist-image, and text tokens in a DROID demonstration (top) and a real-robot rollout (bottom). Feature activations are measured by replaying the recordings through $\pi_{0.5}$-DROID.}
    \label{fig:data_recovery_feature}
  \end{figure}

\begin{wraptable}{r}{0.43\linewidth}
      \vspace{-4mm}
      \centering
      \footnotesize
      \setlength{\tabcolsep}{3pt}
      \resizebox{\linewidth}{!}{%
      \begin{tabular}{@{}lr|rr@{}}
          \toprule
          Method &
          \shortstack{Reopen\\(s) $\uparrow$} &
          \shortstack{Success\\(\%) $\uparrow$} &
          \shortstack{Complete\\(s) $\downarrow$} \\
          \midrule
          Unmodified     & 3.13  & 90  & 12.13 \\
          Random         & 4.34  & 100 & 10.15 \\
          Diff. means    & 5.01  & 65  & 17.49 \\
          kNN            & 3.45  & 100 & 12.82 \\
          Influence      & 5.74  & 100 & 11.12 \\
          Concept (ours) & 12.93 & 100 & 11.79 \\
          \bottomrule
      \end{tabular}%
      }
      \vspace{-1mm}
      \caption{\textbf{Behavior shaping.}
      Mean reopening time (15 perturbed trials) and task performance
      (20 unperturbed trials; 500-step limit).
      Longer reopening indicates stronger recovery suppression;
      completion time averages successful trials only.}
      \label{fig:data_shaping}
      \vspace{-10mm}
\end{wraptable}
\paragraph{Identifying failure recovery features.}
We next ask whether a discovered feature can identify training data responsible for a policy behavior, and whether filtering that data predictably changes the learned policy. We focus on \texttt{F1515} (Figure~\ref{fig:data_recovery_feature}), a three-dimensional BSF feature in layer nine of $\pi_{0.5}$ that activates during \textit{recovery from failed grasps}. We identified it using a small dataset of our own rollouts Figure~\ref{fig:data_recovery_feature} shows the feature's geometry and activations during repeated grasp failures and corrections in a DROID demonstration (top) and a real-robot rollout (bottom). Because it fires broadly across visual tokens, we show mean activation per frame for clarity. The feature activates during missed grasps in both settings and has a distinctly nonlinear, \textit{conical} geometry: multiple SAE features tile this structure, while the BSF captures it in a single block.

\paragraph{From features to training data.}
We identify recovery windows in DROID where \texttt{F1515} activation exceeds a threshold for at least two consecutive evaluated frames (Appendix~\ref{app:data_shaping}), finding failures in approximately 6.5\% of episodes. To test whether these segments teach recovery, we fine-tune from $\pi_{0.5}$-base and mask the loss for any 16-step action chunk overlapping a \texttt{F1515} selected window.  This removes supervision for $1.38\%$ of tokens while preserving observations and trajectory structure.  We compare our BSF-based selection with TRAK attribution~\citep{park2023trak}, a covariance-adjusted difference-of-means labeler, kNN, and random masking, matching  selected-frame budgets and training recipes across  methods. No feature interventions are used at inference time.

\paragraph{Targeted changes in recovery behavior.}
We evaluate recovery by manually closing the gripper before a grasp and measuring the time to reopen it (this remains in distribution because the models have no memory). Across five trials per method (Table~\ref{fig:data_shaping}), concept-based masking increases mean reopening time more than sevenfold, from $1.89$ to $14.25$ seconds, exceeding the mean delay of all comparison methods. In a separate real-world pick-and-place evaluation without induced failures, the concept-filtered policy completes the task with a perfect success rate and similar rollout durations ($11.79$ versus $12.13$ seconds). These results show how interpretable feature-guided data attribution can target specific behaviors while preserving overall task performance in this evaluation.

\section{Conclusion}
\label{sec:discussion}
We have shown that BSFs recover interpretable, multidimensional features in vision-language-action models, providing tools for steering robot behavior and tracing it to training data. Targeted interventions establish these features' behavioral relevance, while feature-guided data masking selectively delays failure recovery and preserves task performance in our evaluation. In terms of limitations, our analysis does not fully explain how these features emerge or participate in policy computation; tracing their geometry to proprioceptive inputs is a promising next step. We demonstrate closed-loop semantic steering, but some interventions are evaluated only open loop. Robust closed-loop visual and geometric steering will require tracking moving image regions and adapting to changing observations and robot state. BSFs are also not uniformly superior: SAEs perform better on some quantitative metrics, and our stronger baselines underscore the importance of unpooled activations and training scale alongside feature geometry.

\newpage
\bibliography{paperpile}

\begin{thebibliography}{53}
\providecommand{\natexlab}[1]{#1}
\providecommand{\url}[1]{\texttt{#1}}
\expandafter\ifx\csname urlstyle\endcsname\relax
  \providecommand{\doi}[1]{doi: #1}\else
  \providecommand{\doi}{doi: \begingroup \urlstyle{rm}\Url}\fi

\bibitem[Agia et~al.(2025)Agia, Sinha, Yang, Antonova, Pavone, Nishimura, Itkina, and Bohg]{agia2025cupid}
Agia, C., Sinha, R., Yang, J., Antonova, R., Pavone, M., Nishimura, H., Itkina, M., and Bohg, J.
\newblock Cupid: Curating data your robot loves with influence functions.
\newblock \emph{arXiv preprint arXiv:2506.19121}, 2025.
\newblock \doi{10.48550/arxiv.2506.19121}.

\bibitem[Bach et~al.(2012)Bach, Jenatton, Mairal, and Obozinski]{bach2012structured}
Bach, F., Jenatton, R., Mairal, J., and Obozinski, G.
\newblock Structured sparsity through convex optimization.
\newblock \emph{Stat. Sci.}, 27\penalty0 (4):\penalty0 450--468, 2012.
\newblock \doi{10.1214/12-sts394}.

\bibitem[Bergen et~al.(2026)Bergen, Bhalla, Baskaran, Loeffler, Sarfati, Gala, Panwar, Aranguri, Fel, Geiger, Kowal, Boppana, Balsam, Lewis, Merullo, McGrath, and Lubana]{bergen2026anatomy}
Bergen, L., Bhalla, U., Baskaran, S., Loeffler, M., Sarfati, R., Gala, D., Panwar, R., Aranguri, S., Fel, T., Geiger, A., Kowal, M., Boppana, S., Balsam, D., Lewis, O., Merullo, J., McGrath, T., and Lubana, E.~S.
\newblock Anatomy of post-training: Using interpretability to characterize data and shape the learning signal.
\newblock \emph{arXiv preprint arXiv:2606.12360}, 2026.
\newblock \doi{10.48550/arxiv.2606.12360}.

\bibitem[Bhalla et~al.(2026)Bhalla, Fel, Rager, Feucht, Haklay, Wurgaft, Boppana, Kowal, Shyam, Merullo, Geiger, and Lubana]{bhalla2026sparse}
Bhalla, U., Fel, T., Rager, C., Feucht, S., Haklay, T., Wurgaft, D., Boppana, S., Kowal, M., Shyam, V., Merullo, J., Geiger, A., and Lubana, E.~S.
\newblock Do sparse autoencoders capture concept manifolds?
\newblock \emph{arXiv preprint arXiv:2604.28119}, 2026.
\newblock \doi{10.48550/arxiv.2604.28119}.

\bibitem[Bhardwaj et~al.(2026)Bhardwaj, Duan, Dan, Ma, and Culbertson]{bhardwaj2026decoding}
Bhardwaj, A., Duan, E.~W., Dan, P., Ma, W.-C., and Culbertson, P.
\newblock Decoding task progress from vla representations.
\newblock \emph{arXiv preprint arXiv:2608.13474}, 2026.
\newblock \doi{10.48550/arxiv.2608.13474}.

\bibitem[Black et~al.(2024)Black, Brown, Driess, Esmail, Equi, Finn, Fusai, Groom, Hausman, Ichter, Jakubczak, Jones, Ke, Levine, Li-Bell, Mothukuri, Nair, Pertsch, Shi, Tanner, Vuong, Walling, Wang, and Zhilinsky]{black2024pi0}
Black, K., Brown, N., Driess, D., Esmail, A., Equi, M., Finn, C., Fusai, N., Groom, L., Hausman, K., Ichter, B., Jakubczak, S., Jones, T., Ke, L., Levine, S., Li-Bell, A., Mothukuri, M., Nair, S., Pertsch, K., Shi, L.~X., Tanner, J., Vuong, Q., Walling, A., Wang, H., and Zhilinsky, U.
\newblock $\pi_0$: A vision-language-action flow model for general robot control.
\newblock \emph{arXiv preprint arXiv:2410.24164}, 2024.
\newblock \doi{10.48550/arxiv.2410.24164}.

\bibitem[Black et~al.(2025)Black, Brown, Darpinian, Dhabalia, Driess, Esmail, Equi, Finn, Fusai, and {others}]{intelligence2025pi05}
Black, K., Brown, N., Darpinian, J., Dhabalia, K., Driess, D., Esmail, A., Equi, M., Finn, C., Fusai, N., and {others}.
\newblock $\pi_{0.5}$: A vision-language-action model with open-world generalization.
\newblock \emph{arXiv preprint arXiv:2504.16054}, 2025.

\bibitem[Bricken et~al.(2023)Bricken, Templeton, Batson, Chen, Jermyn, Conerly, Turner, Anil, Denison, Askell, Lasenby, Wu, Kravec, Schiefer, Maxwell, Joseph, Hatfield-Dodds, Tamkin, Nguyen, McLean, Burke, Hume, Carter, Henighan, and Olah]{bricken2023monosemanticity}
Bricken, T., Templeton, A., Batson, J., Chen, B., Jermyn, A., Conerly, T., Turner, N., Anil, C., Denison, C., Askell, A., Lasenby, R., Wu, Y., Kravec, S., Schiefer, N., Maxwell, T., Joseph, N., Hatfield-Dodds, Z., Tamkin, A., Nguyen, K., McLean, B., Burke, J.~E., Hume, T., Carter, S., Henighan, T., and Olah, C.
\newblock Towards monosemanticity: Decomposing language models with dictionary learning.
\newblock \emph{Transformer Circuits Thread}, 2023.

\bibitem[Brohan et~al.(2023)Brohan, Brown, Carbajal, Chebotar, Choromanski, Ding, Driess, Dubey, Finn, Florence, Fu, Arenas, Gopalakrishnan, Han, Hausman, Herzog, Hsu, Ichter, Irpan, Joshi, Julian, Kalashnikov, Kuang, Leal, Levine, Michalewski, Mordatch, Pertsch, Rao, Reymann, Ryoo, Salazar, Sanketi, Sermanet, Singh, Singh, Soricut, Tran, Vanhoucke, Vuong, Wahid, Welker, Wohlhart, Xiao, Yu, and Zitkovich]{brohan2023rt}
Brohan, A., Brown, N., Carbajal, J., Chebotar, Y., Choromanski, K., Ding, T., Driess, D., Dubey, K.~A., Finn, C., Florence, P.~R., Fu, C., Arenas, M.~G., Gopalakrishnan, K., Han, K., Hausman, K., Herzog, A., Hsu, J., Ichter, B., Irpan, A., Joshi, N.~J., Julian, R.~C., Kalashnikov, D., Kuang, Y., Leal, I., Levine, S., Michalewski, H., Mordatch, I., Pertsch, K., Rao, K., Reymann, K., Ryoo, M., Salazar, G., Sanketi, P.~R., Sermanet, P., Singh, J., Singh, A., Soricut, R., Tran, H., Vanhoucke, V., Vuong, Q., Wahid, A., Welker, S., Wohlhart, P., Xiao, T., Yu, T.-H., and Zitkovich, B.
\newblock Rt-2: Vision-language-action models transfer web knowledge to robotic control.
\newblock \emph{arXiv preprint arXiv:2307.15818}, 2023.
\newblock \doi{10.48550/arxiv.2307.15818}.

\bibitem[Bussmann et~al.(2024)Bussmann, Leask, and Nanda]{bussmann2024batchtopk}
Bussmann, B., Leask, P., and Nanda, N.
\newblock Batchtopk sparse autoencoders.
\newblock \emph{arXiv preprint arXiv:2412.06410}, 2024.
\newblock \doi{10.48550/arxiv.2412.06410}.

\bibitem[Bussmann et~al.(2025)Bussmann, Nabeshima, Karvonen, and Nanda]{bussmann2025learning}
Bussmann, B., Nabeshima, N., Karvonen, A., and Nanda, N.
\newblock Learning multi-level features with matryoshka sparse autoencoders.
\newblock \emph{arXiv preprint arXiv:2503.17547}, 2025.
\newblock \doi{10.48550/arxiv.2503.17547}.

\bibitem[Buurmeijer et~al.(2026)Buurmeijer, Alonso, Swann, and Pavone]{buurmeijer2026observing}
Buurmeijer, H., Alonso, C.~A., Swann, A., and Pavone, M.
\newblock Observing and controlling features in vision-language-action models.
\newblock \emph{arXiv preprint arXiv:2603.05487}, 2026.
\newblock \doi{10.48550/arxiv.2603.05487}.

\bibitem[Cohen et~al.(2020)Cohen, Chung, Lee, and Sompolinsky]{cohen2020separability}
Cohen, U., Chung, S., Lee, D.~D., and Sompolinsky, H.
\newblock Separability and geometry of object manifolds in deep neural networks.
\newblock \emph{Nature communications}, 11\penalty0 (1):\penalty0 746, 2020.
\newblock \doi{10.1038/s41467-020-14578-5}.

\bibitem[Dass et~al.(2026)Dass, Khaddaj, Engstrom, Madry, Ilyas, and Martin-Martin]{dass2026datamil}
Dass, S., Khaddaj, A., Engstrom, L., Madry, A., Ilyas, A., and Martin-Martin, R.
\newblock Datamil: Selecting data for robot imitation learning with datamodels.
\newblock In \emph{International Conference on Learning Representations}, pp.\  1311--1336, 2026.
\newblock \doi{10.48550/arxiv.2505.09603}.

\bibitem[Deshpande et~al.(2026)Deshpande, Guru, Hendrix, Jauhri, Eftekhar, Tripathi, Argus, Salvador, Fang, Wallingford, Pumacay, Kim, Pfeifer, Lee, Wolters, Rayyan, Zhang, Duan, Farley, Han, Vanderbilt, Fox, Farhadi, Chalvatzaki, Shah, and Krishna]{deshpande2026molmob0t}
Deshpande, A., Guru, M., Hendrix, R., Jauhri, S., Eftekhar, A., Tripathi, R., Argus, M., Salvador, J., Fang, H., Wallingford, M., Pumacay, W., Kim, Y., Pfeifer, Q., Lee, Y.-C., Wolters, P., Rayyan, O., Zhang, M., Duan, J., Farley, K., Han, W., Vanderbilt, E., Fox, D., Farhadi, A., Chalvatzaki, G., Shah, D., and Krishna, R.
\newblock Molmob0t: Large-scale simulation enables zero-shot manipulation.
\newblock \emph{arXiv preprint arXiv:2603.16861}, 2026.
\newblock \doi{10.48550/arxiv.2603.16861}.

\bibitem[Eldar \& B{\"{o}}lcskei(2009)Eldar and B{\"{o}}lcskei]{eldar2009block}
Eldar, Y.~C. and B{\"{o}}lcskei, H.
\newblock Block-sparsity: Coherence and efficient recovery.
\newblock In \emph{IEEE International Conference on Acoustics, Speech, and Signal Processing}, pp.\  2885--2888, 2009.
\newblock \doi{10.1109/icassp.2009.4960226}.

\bibitem[Elhage et~al.(2022)Elhage, Hume, Olsson, Schiefer, Henighan, Kravec, Hatfield-Dodds, Lasenby, Drain, Chen, Grosse, McCandlish, Kaplan, Amodei, Wattenberg, and Olah]{elhage2022toy}
Elhage, N., Hume, T., Olsson, C., Schiefer, N., Henighan, T., Kravec, S., Hatfield-Dodds, Z., Lasenby, R., Drain, D., Chen, C., Grosse, R., McCandlish, S., Kaplan, J., Amodei, D., Wattenberg, M., and Olah, C.
\newblock Toy models of superposition.
\newblock \emph{arXiv preprint arXiv:2209.10652}, 2022.
\newblock \doi{10.48550/arxiv.2209.10652}.

\bibitem[Engels et~al.(2025)Engels, Michaud, Liao, Gurnee, and Tegmark]{engels2025not}
Engels, J., Michaud, E.~J., Liao, I., Gurnee, W., and Tegmark, M.
\newblock Not all language model features are one-dimensionally linear.
\newblock In \emph{International Conference on Learning Representations}, volume 2025, pp.\  84591--84622, 2025.
\newblock \doi{10.48550/arxiv.2405.14860}.

\bibitem[Fang et~al.(2026)Fang, Duan, Clay, Wang, Liu, Huang, Fan, Tsai, Chen, Wang, Xing, Cho, Park, Eftekhar, Sushko, Farley, Wadhwa, Harrison, Han, Lee, VanderBilt, Hendrix, Ellawela, Ngoo, Chai, Ren, Farhadi, Fox, and Krishna]{fang2026molmoact2}
Fang, H., Duan, J., Clay, D., Wang, S., Liu, S., Huang, W., Fan, X., Tsai, W.-C., Chen, S., Wang, Y.~R., Xing, S., Cho, J., Park, J.~S., Eftekhar, A., Sushko, P., Farley, K., Wadhwa, A., Harrison, C., Han, W., Lee, Y.-C., VanderBilt, E., Hendrix, R., Ellawela, S., Ngoo, L., Chai, J., Ren, Z., Farhadi, A., Fox, D., and Krishna, R.
\newblock Molmoact2: Action reasoning models for real-world deployment.
\newblock \emph{arXiv preprint arXiv:2605.02881}, 2026.
\newblock \doi{10.48550/arxiv.2605.02881}.

\bibitem[Fel et~al.(2025)Fel, Lubana, Prince, Kowal, Boutin, Papadimitriou, Wang, Wattenberg, Ba, and Konkle]{fel2025archetypal}
Fel, T., Lubana, E., Prince, J.~S., Kowal, M., Boutin, V., Papadimitriou, I., Wang, B., Wattenberg, M., Ba, D.~E., and Konkle, T.
\newblock Archetypal sae: Adaptive and stable dictionary learning for concept extraction in large vision models.
\newblock \emph{arXiv preprint arXiv:2502.12892}, 2025.
\newblock \doi{10.48550/arxiv.2502.12892}.

\bibitem[Fel et~al.(2026{\natexlab{a}})Fel, Kowal, Jacobs, Hazra, Bhalla, Sharkey, Bushnaq, Grant, Haklay, Icard, Rager, Pearce, Wurgaft, Swann, Doshi, Boppana, Tigges, Cammarata, Serre, Shyam, Lewis, McGrath, Merullo, Lubana, and Geiger]{fel2026bsf}
Fel, T., Kowal, M., Jacobs, M., Hazra, D., Bhalla, U., Sharkey, L., Bushnaq, L., Grant, S., Haklay, T., Icard, T., Rager, C., Pearce, M., Wurgaft, D., Swann, A., Doshi, F., Boppana, S., Tigges, C., Cammarata, N., Serre, T., Shyam, V., Lewis, O., McGrath, T., Merullo, J., Lubana, E.~S., and Geiger, A.
\newblock Structuring sparsity: Block-sparse featurizers capture visual concept manifolds.
\newblock \emph{arXiv preprint arXiv:2606.25234}, 2026{\natexlab{a}}.
\newblock \doi{10.48550/arxiv.2606.25234}.

\bibitem[Fel et~al.(2026{\natexlab{b}})Fel, Wang, Lepori, Kowal, Lee, Balestriero, Joseph, Lubana, Konkle, Ba, and Wattenberg]{fel2026into}
Fel, T., Wang, B., Lepori, M.~A., Kowal, M., Lee, A., Balestriero, R., Joseph, S., Lubana, E., Konkle, T., Ba, D.~E., and Wattenberg, M.
\newblock Into the rabbit hull: From task-relevant concepts in dino to minkowski geometry.
\newblock In \emph{International Conference on Learning Representations}, pp.\  154538--154584, 2026{\natexlab{b}}.
\newblock \doi{10.48550/arxiv.2510.08638}.

\bibitem[Gallifant et~al.(2025)Gallifant, Chen, Sasse, Aerts, Hartvigsen, and Bitterman]{gallifant2025sparse}
Gallifant, J., Chen, S., Sasse, K., Aerts, H. J. W.~L., Hartvigsen, T., and Bitterman, D.
\newblock Sparse autoencoder features for classifications and transferability.
\newblock In \emph{Conference on Empirical Methods in Natural Language Processing}, pp.\  29939--29963, 2025.
\newblock \doi{10.48550/arxiv.2502.11367}.

\bibitem[Gao et~al.(2025)Gao, Tour, Tillman, Goh, Troll, Radford, Sutskever, Leike, and Wu]{gao2025scaling}
Gao, L., Tour, T. D.~l., Tillman, H., Goh, G., Troll, R., Radford, A., Sutskever, I., Leike, J., and Wu, J.
\newblock Scaling and evaluating sparse autoencoders.
\newblock In \emph{International Conference on Learning Representations}, pp.\  26721--26754, 2025.
\newblock \doi{10.48550/arxiv.2406.04093}.

\bibitem[Grant et~al.(2026)Grant, Zhao, and Wang]{grant2026not}
Grant, B., Zhao, X., and Wang, P.
\newblock Not all features are created equal: A mechanistic study of vision-language-action models.
\newblock \emph{arXiv preprint arXiv:2603.19233}, 2026.
\newblock \doi{10.48550/arxiv.2603.19233}.

\bibitem[H{\"{a}}on et~al.(2025)H{\"{a}}on, Stocking, Chuang, and Tomlin]{haon2025mechanistic}
H{\"{a}}on, B., Stocking, K., Chuang, I., and Tomlin, C.
\newblock Mechanistic interpretability for steering vision-language-action models.
\newblock \emph{arXiv preprint arXiv:2509.00328}, 2025.
\newblock \doi{10.48550/arxiv.2509.00328}.

\bibitem[Hejna et~al.(2025)Hejna, Mirchandani, Balakrishna, Xie, Wahid, Tompson, Sanketi, Shah, Devin, and Sadigh]{hejna2025robot}
Hejna, J., Mirchandani, S., Balakrishna, A., Xie, A., Wahid, A., Tompson, J., Sanketi, P., Shah, D., Devin, C., and Sadigh, D.
\newblock Robot data curation with mutual information estimators.
\newblock \emph{arXiv preprint arXiv:2502.08623}, 2025.
\newblock \doi{10.48550/arxiv.2502.08623}.

\bibitem[Hindupur et~al.(2025)Hindupur, Lubana, Fel, and Ba]{hindupur2026projecting}
Hindupur, S. S.~R., Lubana, E., Fel, T., and Ba, D.~E.
\newblock Projecting assumptions: The duality between sparse autoencoders and concept geometry.
\newblock \emph{Advances in Neural Information Processing Systems}, pp.\  11649--11699, 2025.
\newblock \doi{10.48550/arxiv.2503.01822}.

\bibitem[Huben et~al.(2024)Huben, Cunningham, Smith, Ewart, and Sharkey]{huben2024sparse}
Huben, R., Cunningham, H., Smith, L., Ewart, A., and Sharkey, L.
\newblock Sparse autoencoders find highly interpretable features in language models.
\newblock In \emph{International Conference on Learning Representations}, volume 2024, pp.\  7827--7845, 2024.

\bibitem[Jenatton et~al.(2010)Jenatton, Obozinski, and Bach]{jenatton2010structured}
Jenatton, R., Obozinski, G., and Bach, F.
\newblock Structured sparse principal component analysis.
\newblock In \emph{International Conference on Artificial Intelligence and Statistics}, pp.\  366--373, 2010.
\newblock \doi{10.48550/arxiv.0909.1440}.

\bibitem[Jin et~al.(2026)Jin, Chatterjee, Kumar, and Paleja]{jin2026event}
Jin, X., Chatterjee, A., Kumar, P., and Paleja, R.
\newblock Event-grounded sparse autoencoders for vision-language-action policies.
\newblock \emph{arXiv preprint arXiv:2605.17204}, 2026.
\newblock \doi{10.48550/arxiv.2605.17204}.

\bibitem[Khan et~al.(2025)Khan, Boskov, Anwar, and Khan]{khan2025controlling}
Khan, M.~A., Boskov, N., Anwar, F.~M., and Khan, M.~A.
\newblock Controlling vision--language--action policies through sparse latent directions.
\newblock In \emph{Mechanistic Interpretability Workshop at NeurIPS 2025}, 2025.

\bibitem[Khazatsky et~al.(2024)Khazatsky, Pertsch, Nair, Balakrishna, Dasari, Karamcheti, Nasiriany, Srirama, Chen, Ellis, Fagan, Hejna, Itkina, Lepert, Ma, Miller, Wu, Belkhale, Dass, Ha, Jain, Lee, Lee, Memmel, Park, Radosavovic, Wang, Zhan, Black, Chi, Hatch, Lin, Lu, Mercat, Rehman, Sanketi, Sharma, Simpson, Vuong, Walke, Wulfe, Xiao, Yang, Yavary, Zhao, Agia, Baijal, Castro, Chen, Chen, Chung, Drake, Foster, Gao, Herrera, Heo, Hsu, Hu, Irshad, Jackson, Le, Li, Lin, Lin, Ma, Maddukuri, Mirchandani, Morton, Nguyen, O'Neill, Scalise, Seale, Son, Tian, Tran, Wang, Wu, Xie, Yang, Yin, Zhang, Bastani, Berseth, Bohg, Goldberg, Gupta, Gupta, Jayaraman, Lim, Malik, Martin-Martin, Ramamoorthy, Sadigh, Song, Wu, Yip, Zhu, Kollar, Levine, and Finn]{khazatsky2024droid}
Khazatsky, A., Pertsch, K., Nair, S., Balakrishna, A., Dasari, S., Karamcheti, S., Nasiriany, S., Srirama, M.~K., Chen, L.~Y., Ellis, K., Fagan, P., Hejna, J., Itkina, M., Lepert, M., Ma, Y., Miller, P., Wu, J., Belkhale, S., Dass, S., Ha, H., Jain, A., Lee, A., Lee, Y., Memmel, M., Park, S., Radosavovic, I., Wang, K., Zhan, A., Black, K., Chi, C., Hatch, K., Lin, S., Lu, J., Mercat, J.-P., Rehman, A., Sanketi, P.~R., Sharma, A., Simpson, C., Vuong, Q., Walke, H., Wulfe, B., Xiao, T., Yang, J., Yavary, A., Zhao, T., Agia, C., Baijal, R., Castro, M.~G., Chen, D., Chen, Q., Chung, T., Drake, J., Foster, E.~P., Gao, J., Herrera, D., Heo, M., Hsu, K., Hu, J., Irshad, M.~Z., Jackson, D., Le, C., Li, Y., Lin, K., Lin, R., Ma, Z.-H., Maddukuri, A., Mirchandani, S., Morton, D., Nguyen, T., O'Neill, A., Scalise, R., Seale, D., Son, V., Tian, S., Tran, E., Wang, A., Wu, Y., Xie, A., Yang, J.-Y., Yin, P., Zhang, Y., Bastani, O., Berseth, G., Bohg, J., Goldberg, K., Gupta, A., Gupta, A., Jayaraman, D., Lim, J.~J., Malik,
  J., Martin-Martin, R., Ramamoorthy, S., Sadigh, D., Song, S., Wu, J.-J., Yip, M.~C., Zhu, Y.-K., Kollar, T., Levine, S., and Finn, C.
\newblock Droid: A large-scale in-the-wild robot manipulation dataset.
\newblock \emph{arXiv preprint arXiv:2403.12945}, 2024.
\newblock \doi{10.48550/arxiv.2403.12945}.

\bibitem[Kim et~al.(2024)Kim, Pertsch, Karamcheti, Xiao, Balakrishna, Nair, Rafailov, Foster, Lam, Sanketi, Vuong, Kollar, Burchfiel, Tedrake, Sadigh, Levine, Liang, and Finn]{kim2024openvla}
Kim, M.~J., Pertsch, K., Karamcheti, S., Xiao, T., Balakrishna, A., Nair, S., Rafailov, R., Foster, E., Lam, G., Sanketi, P.~R., Vuong, Q., Kollar, T., Burchfiel, B., Tedrake, R., Sadigh, D., Levine, S., Liang, P., and Finn, C.
\newblock Openvla: An open-source vision-language-action model.
\newblock \emph{arXiv preprint arXiv:2406.09246}, 2024.
\newblock \doi{10.48550/arxiv.2406.09246}.

\bibitem[Koh \& Liang(2017)Koh and Liang]{koh2017understanding}
Koh, P.~W. and Liang, P.
\newblock Understanding black-box predictions via influence functions.
\newblock In \emph{International conference on machine learning}, pp.\  1885--1894, 2017.
\newblock \doi{10.48550/arxiv.1703.04730}.

\bibitem[Kowal et~al.(2026)Kowal, Paulo, Jaburi, Tseng, McKinney, Heimersheim, Tucker, Gleave, and Pelrine]{kowal2026concept}
Kowal, M., Paulo, G., Jaburi, L., Tseng, T., McKinney, L.~E., Heimersheim, S., Tucker, A.~D., Gleave, A., and Pelrine, K.
\newblock Concept influence: Leveraging interpretability to improve performance and efficiency in training data attribution.
\newblock \emph{arXiv preprint arXiv:2602.14869}, 2026.
\newblock \doi{10.48550/arxiv.2602.14869}.

\bibitem[Kvinge et~al.(2024)Kvinge, Jorgenson, Brown, Godfrey, and Emerson]{kvinge2024internal}
Kvinge, H., Jorgenson, G., Brown, D., Godfrey, C., and Emerson, T.
\newblock Internal representations of vision models through the lens of frames on data manifolds.
\newblock In Sanborn, S., Shewmake, C., Azeglio, S., and Miolane, N. (eds.), \emph{Proceedings of the 2nd NeurIPS Workshop on Symmetry and Geometry in Neural Representations}, volume 228 of \emph{Proceedings of Machine Learning Research}, pp.\  75--115. PMLR, 2024.
\newblock \doi{10.48550/arxiv.2211.10558}.
\newblock URL \url{https://proceedings.mlr.press/v228/kvinge24a.html}.

\bibitem[Lee et~al.(2026)Lee, Duan, Fang, Deng, Liu, Li, Fang, Zhang, Wang, Lee, Han, Pumacay, Wu, Hendrix, Farley, VanderBilt, Farhadi, Fox, and Krishna]{lee2026molmoact}
Lee, J., Duan, J., Fang, H., Deng, Y.-Q., Liu, S., Li, B.-Y., Fang, B., Zhang, J.-Y., Wang, Y.-R., Lee, S.-H., Han, W., Pumacay, W., Wu, A., Hendrix, R., Farley, K., VanderBilt, E., Farhadi, A., Fox, D., and Krishna, R.
\newblock Molmoact: Action reasoning models that can reason in space.
\newblock In \emph{IEEE International Conference on Robotics and Automation}, pp.\  2585--2592, 2026.
\newblock \doi{10.1109/icra57385.2026.11696546}.

\bibitem[Marks et~al.(2025)Marks, Treutlein, Bricken, Lindsey, Marcus, Mishra-Sharma, Ziegler, Ameisen, Batson, Belonax, Bowman, Carter, Chen, Cunningham, Denison, Dietz, Golechha, Khan, Kirchner, Leike, Meek, Nishimura-Gasparian, Ong, Olah, Pearce, Roger, Salle, Shih, Tong, Thomas, Rivoire, Jermyn, MacDiarmid, Henighan, and Hubinger]{marks2025auditing}
Marks, S., Treutlein, J., Bricken, T., Lindsey, J., Marcus, J., Mishra-Sharma, S., Ziegler, D., Ameisen, E., Batson, J., Belonax, T., Bowman, S.~R., Carter, S., Chen, B., Cunningham, H., Denison, C., Dietz, F., Golechha, S., Khan, A., Kirchner, J., Leike, J., Meek, A., Nishimura-Gasparian, K., Ong, E., Olah, C., Pearce, A., Roger, F., Salle, J., Shih, A., Tong, M., Thomas, D., Rivoire, K., Jermyn, A., MacDiarmid, M., Henighan, T., and Hubinger, E.
\newblock Auditing language models for hidden objectives.
\newblock \emph{arXiv preprint arXiv:2503.10965}, 2025.
\newblock \doi{10.48550/arxiv.2503.10965}.

\bibitem[Michaud et~al.(2025)Michaud, Gorton, and McGrath]{michaud2025understanding}
Michaud, E.~J., Gorton, L., and McGrath, T.
\newblock Understanding sparse autoencoder scaling in the presence of feature manifolds.
\newblock \emph{arXiv preprint arXiv:2509.02565}, 2025.
\newblock \doi{10.48550/arxiv.2509.02565}.

\bibitem[Molinari et~al.(2025)Molinari, Nevali, Navani, and Younis]{molinari2025emergent}
Molinari, M., Nevali, L., Navani, S., and Younis, O.~G.
\newblock Emergent world representations in openvla.
\newblock \emph{arXiv preprint arXiv:2509.24559}, 2025.
\newblock \doi{10.48550/arxiv.2509.24559}.

\bibitem[Olson et~al.(2025)Olson, Hinck, Ratzlaff, Li, Howard, Lal, and Tseng]{olson2025probing}
Olson, M., Hinck, M., Ratzlaff, N., Li, C.-B., Howard, P., Lal, V., and Tseng, S.-Y.
\newblock Probing the representational power of sparse autoencoders in vision models.
\newblock In \emph{Proceedings of the IEEE/CVF International Conference on Computer Vision}, pp.\  6226--6236, 2025.
\newblock \doi{10.1109/iccvw69036.2025.00648}.

\bibitem[Park et~al.(2023{\natexlab{a}})Park, Choe, and Veitch]{park2023linear}
Park, K., Choe, Y.~J., and Veitch, V.
\newblock The linear representation hypothesis and the geometry of large language models.
\newblock \emph{arXiv preprint arXiv:2311.03658}, pp.\  39643--39666, 2023{\natexlab{a}}.
\newblock \doi{10.48550/arxiv.2311.03658}.

\bibitem[Park et~al.(2023{\natexlab{b}})Park, Georgiev, Ilyas, Leclerc, and M{\c{a}}dry]{park2023trak}
Park, S.~M., Georgiev, K., Ilyas, A., Leclerc, G., and M{\c{a}}dry, A.
\newblock Trak: Attributing model behavior at scale.
\newblock \emph{arXiv preprint arXiv:2303.14186}, pp.\  27074--27113, 2023{\natexlab{b}}.
\newblock \doi{10.48550/arxiv.2303.14186}.

\bibitem[Prasad et~al.(2026)Prasad, Watts, Merullo, Gala, Lewis, McGrath, and Lubana]{prasad2026features}
Prasad, A.~V., Watts, C., Merullo, J., Gala, D., Lewis, O., McGrath, T., and Lubana, E.~S.
\newblock Features as rewards: Scalable supervision for open-ended tasks via interpretability.
\newblock \emph{arXiv preprint arXiv:2602.10067}, 2026.
\newblock \doi{10.48550/arxiv.2602.10067}.

\bibitem[Rajamanoharan et~al.(2024)Rajamanoharan, Lieberum, Sonnerat, Conmy, Varma, Kram\'{a}r, and Nanda]{rajamanoharan2024jumping}
Rajamanoharan, S., Lieberum, T., Sonnerat, N., Conmy, A., Varma, V., Kram\'{a}r, J., and Nanda, N.
\newblock Jumping ahead: Improving reconstruction fidelity with jumprelu sparse autoencoders.
\newblock \emph{arXiv preprint arXiv:2407.14435}, 2024.
\newblock \doi{10.48550/arxiv.2407.14435}.

\bibitem[Rimsky et~al.(2024)Rimsky, Gabrieli, Schulz, Tong, Hubinger, and Turner]{rimsky2024steering}
Rimsky, N., Gabrieli, N., Schulz, J., Tong, M., Hubinger, E., and Turner, A.
\newblock Steering llama 2 via contrastive activation addition.
\newblock In \emph{Proceedings of the 62nd Annual Meeting of the Association for Computational Linguistics (Volume 1: Long Papers)}, pp.\  15504--15522, 2024.

\bibitem[Sim\'{e}oni et~al.(2025)Sim\'{e}oni, Vo, Seitzer, Baldassarre, Oquab, Jose, Khalidov, Szafraniec, Yi, Ramamonjisoa, Massa, Haziza, Wehrstedt, Wang, Darcet, Moutakanni, Sentana, Roberts, Vedaldi, Tolan, Brandt, Couprie, Mairal, J\'{e}gou, Labatut, and Bojanowski]{simeoni2025dinov3}
Sim\'{e}oni, O., Vo, H.~V., Seitzer, M., Baldassarre, F., Oquab, M., Jose, C., Khalidov, V., Szafraniec, M., Yi, S., Ramamonjisoa, M., Massa, F., Haziza, D., Wehrstedt, L., Wang, J., Darcet, T., Moutakanni, T., Sentana, L., Roberts, C., Vedaldi, A., Tolan, J., Brandt, J., Couprie, C., Mairal, J., J\'{e}gou, H., Labatut, P., and Bojanowski, P.
\newblock Dinov3.
\newblock \emph{arXiv preprint arXiv:2508.10104}, 2025.
\newblock \doi{10.48550/arxiv.2508.10104}.

\bibitem[Swann et~al.(2026)Swann, McGranahan, Buurmeijer, Kennedy, and Schwager]{swann2026sparse}
Swann, A., McGranahan, L., Buurmeijer, H., Kennedy, III, M., and Schwager, M.
\newblock Sparse autoencoders reveal interpretable and steerable features in vla models.
\newblock \emph{arXiv preprint arXiv:2603.19183}, 2026.
\newblock \doi{10.48550/arxiv.2603.19183}.

\bibitem[Templeton et~al.(2026)Templeton, Conerly, Marcus, Lindsey, Bricken, Chen, Pearce, Citro, Ameisen, Jones, Cunningham, Turner, McDougall, MacDiarmid, Tamkin, Durmus, Hume, Mosconi, Daniel~Freeman, Sumers, Rees, Batson, Jermyn, Carter, Olah, and Henighan]{templeton2024scaling}
Templeton, A., Conerly, T., Marcus, J., Lindsey, J., Bricken, T., Chen, B., Pearce, A., Citro, C., Ameisen, E., Jones, A., Cunningham, H., Turner, N.~L., McDougall, C., MacDiarmid, M., Tamkin, A., Durmus, E., Hume, T., Mosconi, F., Daniel~Freeman, C., Sumers, T.~R., Rees, E., Batson, J., Jermyn, A., Carter, S., Olah, C., and Henighan, T.
\newblock Scaling monosemanticity: Extracting interpretable features from claude 3 sonnet.
\newblock \emph{arXiv [cs.AI]}, 28~May 2026.
\newblock \doi{10.48550/arxiv.2605.29358}.

\bibitem[Turner et~al.(2023)Turner, Thiergart, Leech, Udell, Vazquez, Mini, and MacDiarmid]{turner2023steering}
Turner, A.~M., Thiergart, L., Leech, G., Udell, D., Vazquez, J.~J., Mini, U., and MacDiarmid, M.
\newblock Steering language models with activation engineering.
\newblock \emph{arXiv preprint arXiv:2308.10248}, 2023.
\newblock \doi{10.48550/arxiv.2308.10248}.

\bibitem[Yuan \& Lin(2006)Yuan and Lin]{yuan2006model}
Yuan, M. and Lin, Y.
\newblock Model selection and estimation in regression with grouped variables.
\newblock \emph{Journal of the Royal Statistical Society Series B: Statistical Methodology}, 68\penalty0 (1):\penalty0 49--67, 2006.

\bibitem[Zhu et~al.(2026)Zhu, Khalili, and Zhu]{zhu2026abstopk}
Zhu, X., Khalili, M., and Zhu, Z.
\newblock Abstopk: Rethinking sparse autoencoders for bidirectional features.
\newblock In \emph{International Conference on Learning Representations}, pp.\  50325--50352, 2026.
\newblock \doi{10.48550/arxiv.2510.00404}.

\end{thebibliography}
\bibliographystyle{bibliostyle}

\newpage
\appendix
\clearpage

\section{Behavior Shaping: Implementation Details}
\label{app:data_shaping}

\paragraph{Concept Based Attribution.}
We labeled 81 timesteps of failed-grasp \emph{recovery} and 25 timesteps of successful grasps, drawn from real-robot rollouts of $\pi_{0.5}$-DROID on a pick-and-place task. \texttt{F1515} fires on 56 of the 81 recovery timesteps and on none of the 25 successful grasps. We use this feature to scan the full DROID training set for other failure-recovery behavior. For each timestep we record the \texttt{F1515} block norm summed over all image and text tokens, $E$, and the number of tokens where the block is active, $C$. A timestep is \emph{active} if $E \geq 150$ and $C \geq 50$. Runs of at least two consecutive active frames, with single-frame gaps bridged, form a recovery event. Across the entire dataset this flags 4{,}866 of 74{,}604 episodes (6.5\%). The median duration is 1.1s. 

\begin{figure}[h]
    \centering
    \includegraphics[width=\linewidth]{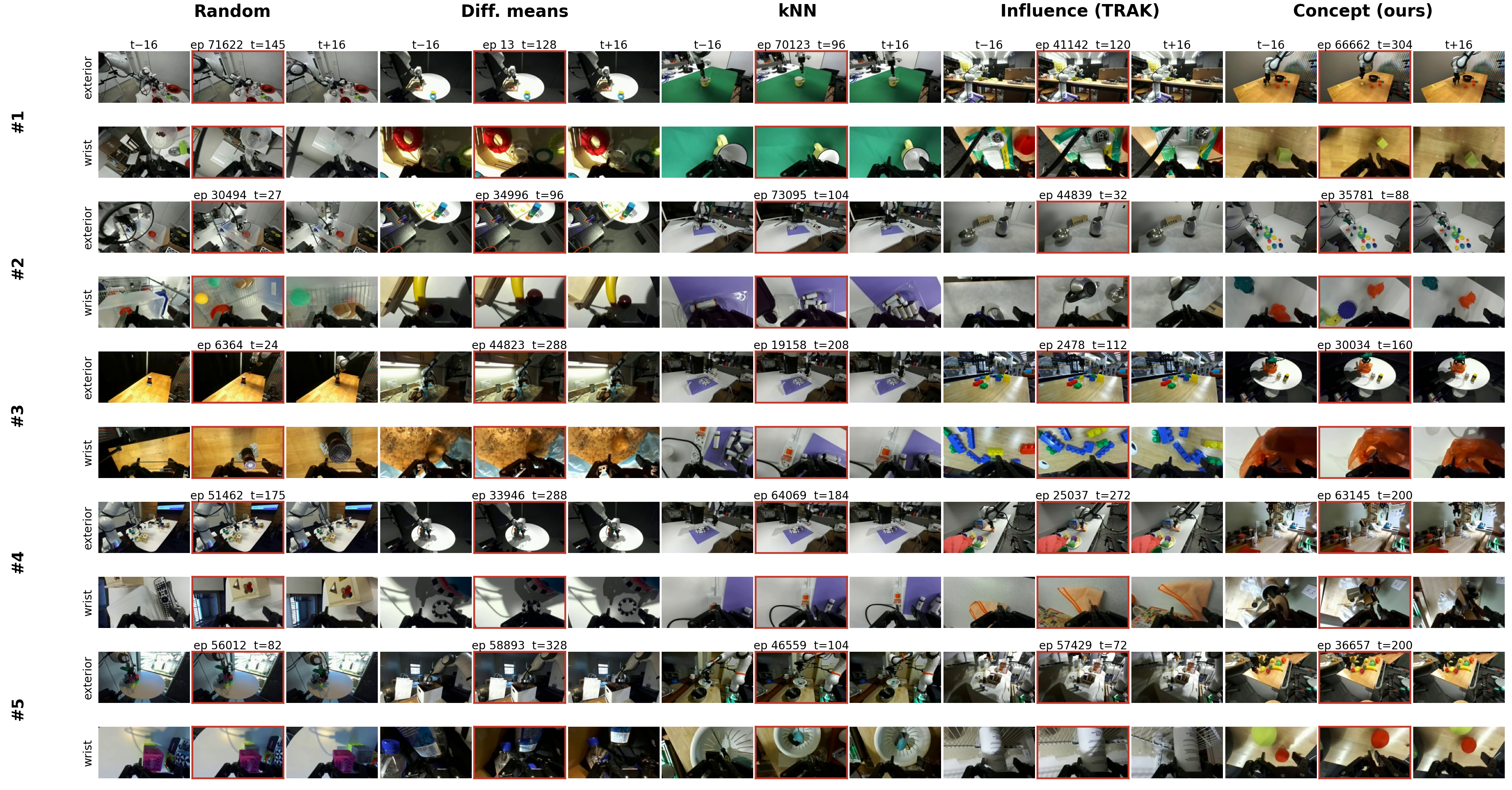}
    \caption{\textbf{Highest-scoring frames for each labeler.} Each column shows five DROID episodes for one method. For the four scored methods, these are the five highest-scoring episodes, and $t$ is the highest-scoring evaluated frame within that method's selected windows. Random has no score, so its column shows five randomly drawn windows with $t$ at each window's midpoint; its row order carries no meaning.}
    \label{fig:top5_labelers}
\end{figure}

\paragraph{Baselines.}
Every baseline scores the same evaluated frames. Its threshold is chosen so that its windows, built with the same event and padding rules, cover exactly as many raw frames as the
\texttt{F1515} windows (150{,}656).
\begin{itemize}
  \item \textbf{Difference of means.} A direction in the mean-pooled layer-9 residual, fit to separate the 81 recovery timesteps from all other frames of the same seven episodes, with the within-class covariance shrunk toward the identity ($\alpha = 0.95$, chosen by nested leave-one-episode-out cross-validation). Frames are scored by projection onto this direction. \item \textbf{kNN.} Each frame's mean-pooled layer-9 residual, centered on the annotated-rollout mean, is scored by its mean cosine similarity to the five nearest recovery timesteps minus its mean cosine similarity to the five nearest successful-grasp timesteps. 
  \item \textbf{Influence Functions.} Following CUPID~\citep{agia2025cupid}, we estimate \emph{action influence}, the effect of a training state--action pair on the policy's log-likelihood of a query action, with TRAK~\citep{park2023trak}. We compute per-example gradients of the flow-matching loss with respect to the action expert (430M parameters) of the baseline policy, randomly project them to 4{,}096 dimensions, and apply the Gauss--Newton kernel. CUPID weights action influence by the binary return of closed-loop rollouts, estimating each demonstration's effect on expected return, and curates whole demonstrations. We target a specific behavior rather than task success. The query is therefore the annotated recovery timesteps, contrasted against the successful-grasp timesteps, and each training frame is scored as its mean influence on the recovery queries minus its mean influence on the successful-grasp queries. Scores stay at the frame level, so the same budget and windowing apply as for every other labeler.
  \item \textbf{Random.} Windows with the same count and length distribution as the \texttt{F1515} windows, placed uniformly at random in episodes \texttt{F1515} did not flag.
\end{itemize}
Selections overlap little with the concept windows: Jaccard similarity over raw frames is 0.20 for difference of means, 0.09 for kNN, 0.02 for TRAK and 0.00 for random. Figure~\ref{fig:top5_labelers} shows the highest-scoring frames selected by each method.

\paragraph{Training.}
All policies are fine-tuned from $\pi_{0.5}$-base on 98\% of DROID (73{,}112 episodes); the remaining 2\% is held out. Training runs for 25k steps at batch size 1024 with a constant learning rate of $10^{-4}$ after a 1k-step warmup, on 32 H100s. All runs share the same seed and data order. For the masked policies, the flow-matching loss is zeroed for any 16-step action chunk containing a selected frame; observations are unchanged. Because window lengths differ, the masked fraction of action tokens varies slightly: 1.38\% for concept, 1.28\% for difference of means, 1.27\% for kNN, 1.44\% for TRAK and 1.27\% for random.

\section{Steering}
\subsection{Open-Loop Steering Protocol}
\label{app:steering_protocol}

\begin{figure}[h]
\centering
\includegraphics[width=\linewidth]{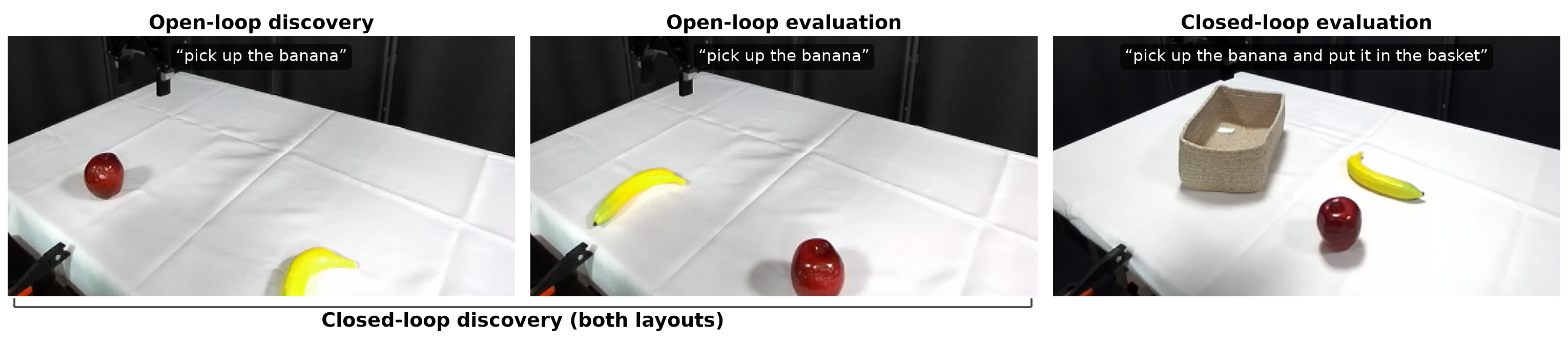}
\caption{\textbf{Open and closed-loop steering setups}. \emph{Open loop}
(\Cref{app:steering_protocol}): we find candidate semantic features and evaluate them on the mirrored scene for semantic features. Visual edits are instead evaluated under the other object's instruction, and geometric edits on held-out object pairs. \emph{Closed loop}: feature sets are discovered on both mirrored layouts, then fixed and tested on new scenes and tasks, ``pick up the banana and put it in the basket".}
\label{fig:steering_scenes}
\end{figure}

Our goal is to find a set of features which steers the robot toward a target object on a set discovery inputs and then evaluated, on held-out inputs with fresh noise seeds. Consider a scene with a banana on the left and an apple on the right. We task the robot with ``pick up the banana", our goal is to steer it to instead pick up the apple using three different means: steering semantic representations (changing what the model thinks the target object is), steering visual tokens (changing where the model sees the target object)  or geometry representations (changing the model's internal representation of the location of the target object). For each of these we have a different held out set to evaluate our steering on. Semantic steering is evaluated on the other mirrored layout. So for example if we have a set of features which redirects the policy from the banana towards the apple we may evaluate this on a different scene where the position of the banana and apple are flipped. For visual token steering we evaluate with the other object's instruction and for geometry we evaluate on an entirely new scene (leave-one-pair-out over the 13 pairs). The experimental setup is shown in \Cref{fig:steering_scenes}. Steering results are conducted on MolmoACT2. 

\paragraph{Finding features.} We want to find a compact set of features that steers the policy as well as patching the full activation does, since the number of features needed measures how directly a featurizer's units capture the representation that drives the behaviour. Searching all feature subsets is intractable, so we use a greedy procedure. Candidates are the BSF blocks or SAE latents are ranked by how much their activation differs between the prompted and redirected conditions. We add features to a group until its steering effect recovers at least 75\% of the patching effect. Then we remove features one at a time while the set still recovers over 75\% of patching effect, so the reported sets are small but not guaranteed to be the smallest possible.

\paragraph{Applying features.} The policy represents the task at different layers and tokens in different forms. Activation patching localizes three sites where the target object is determined: the early noun token encodes \emph{which} object is requested, mid-layer image tokens encode \emph{where} each object appears, and the late action-readout token encodes \emph{where} the arm should move. At the L3 noun token, the selected features' contribution is replaced with that of the redirected instruction. At the 63 L7 image tokens, the recipient's activation within the selected features' subspace is replaced, token by token, with that of the mirrored scene. The 63 tokens are a fixed $9\times7$ window (rows 7--13, columns 2--10 of the $14\times14$ exo-camera patch grid) covering the table region where both objects appear in every layout. These tokens are 78\% of the image-token activation difference between mirrored layouts and the background remains the same between scenes. Geometric edits add one fixed vector at the L23 action-readout token.

\paragraph{Measuring recovery.} We compare the endpoint of the predicted action chunk under steering and under activation patching, each relative to the unmodified policy and averaged over six held-out noise seeds:
\[
R = 100\,\frac{\langle \Delta x_{\mathrm{steer}}, \Delta x_{\mathrm{patch}}\rangle}
{\|\Delta x_{\mathrm{patch}}\|^2}.
\]
A value of 100\% means steering moves the endpoint as far as patching, in the same direction. Our patching replaces the full activation at the noun token, all 196 exo-image tokens, or all 79 text tokens at the intervention layer.

\subsection{Closed-loop Steering}

Feature sets for the closed-loop evaluation are selected with the semantic procedure of Appendix~\ref{app:steering_protocol} and shown in \Cref{fig:steering_combined}. We evaluate on new untested scenes and tasks Because no benchmark scene needs to be held out, each object pair and direction is calibrated on both mirrored layouts at once. Feature sets are fixed before evaluation and applied without gain tuning at the noun token of the prompt ``pick up the [noun] and put it in the basket'' throughout each episode. Note that this is different than the scene and task we used to identify the relevant features: ``pick up the [noun]``. Each object pair is evaluated in both directions with 10 trials per direction.

\begin{table}[h]
\centering
\small
\setlength{\tabcolsep}{3pt}
\begin{tabular}{@{}ll cccc cccc@{}}
\toprule
& & \multicolumn{4}{c}{A $\rightarrow$ B} & \multicolumn{4}{c}{B $\rightarrow$ A} \\
\cmidrule(lr){3-6}\cmidrule(lr){7-10}
Objects (A / B) & Featurizer & \# Feat. & Select & Grasp & Successful & \# Feat. & Select & Grasp & Successful \\
\midrule
Apple / banana & SAE & 2 & 10/10 & 9/10  & 7/10  & 2 & 8/10  & 8/10  & 8/10  \\
               & BSF & 2 & 10/10 & 10/10 & 9/10  & 3 & 10/10 & 9/10  & 9/10  \\
\midrule
Cable / bowl   & SAE & 3 & 10/10 & 9/10  & 8/10  & 2 & 10/10 & 10/10 & 9/10  \\
               & BSF & 2 & 10/10 & 10/10 & 10/10 & 2 & 10/10 & 10/10 & 10/10 \\
\midrule
Towel / mouse  & SAE & 4 & 6/10  & 5/10  & 4/10  & 2 & 9/10  & 9/10  & 9/10  \\
               & BSF & 4 & 10/10 & 9/10  & 8/10  & 1 & 10/10 & 10/10 & 10/10 \\
\bottomrule
\end{tabular}
\caption{\textbf{Closed-loop redirection by direction.} Trials out of 10. Full version of \Cref{tab:steering_rollouts}. A $\rightarrow$ B: instructed to pick up A while steered toward B. Feat.: number of features in the intervention; Grasp counts trials that grasp the redirected object; Successful counts trials that place it in the basket.}
\label{tab:steering_rollouts_full}
\end{table}

\section{Featurizer Training and Evaluation}
\label{app:collection_training}
\label{app:collection}

\paragraph{Activation Collection}
We record the residual stream at the output of language backbone block $L$ (zero-indexed). Featurizers are trained at every odd block of MolmoAct2 ($L\in\{1,3,\ldots,35\}$) and of $\pi_{0.5}$ ($L\in\{1,3,\ldots,17\}$), and at blocks $\{1,3,13,15,24\}$ of MolmoBot. Every image, text, and state token of the is kept as a separate activation. We collect activations from MolmoAct2 and $\pi_{0.5}$ on the DROID dataset~\citep{khazatsky2024droid}.For MolmoBot, which is trained entirely in simulation, we collect activations on its own training distribution \citep{deshpande2026molmob0t}. We collect and train on approximately 50M activations for each model and layer combination. To eliminate issues with outlier tokens we filter tokens by activation magnitude $[p_{0.5},\,\min(p_{99.5},\,8\cdot\mathrm{median})]$, removes between 1--5\% of tokens depending on the layer. We then subtract a per-position mean computed over training tokens and rescale activations to a mean squared norm of $d$. 

\paragraph{Training}
Decoder blocks are initialized as Gaussian matrices with unit Frobenius norm, the encoder is initialized to the decoder transpose, and $\bm{b}_{\mathrm{enc}}=\bm{0}$. Once training begins the encoder and decoder are untied. The loss is the squared error of reconstructing the clean input from a lightly corrupted one: before encoding we add isotropic Gaussian noise with per-coordinate standard deviation $0.1\|\bm{x}\|_2/\sqrt{d}$ (1\% of the token's energy). To revive unused blocks we add the AuxK loss of~\citet{gao2025scaling} The featurizers used throughout the paper (phenomenology, steering, data shaping, and the per-layer comparison) are trained with a $\blocksize=3$ BSF with $\topk=21$ and the SAEs are a $\blocksize=1$ BSF with $\topk=63$. Both have the same number of coordinates in total. They train for 10 epochs, on one H200. We use Adam with a constant learning rate of $10^{-4}$ on batches of 16{,}384 tokens drawn from an 8M-token shuffle buffer; BatchTopK selects blocks during training and a fixed threshold (an EMA of the training thresholds, decay $0.99$) at inference, a block counts as dead for AuxK (weight $1/\topk$) after 1,000 steps without selection, and every 20th episode is held out for evaluation (at most 1M tokens).

\paragraph{Fisher-weighted featurizers}
\label{app:fisher}
Section~\ref{sec:preliminaries} contrasts the standard reconstruction loss (Eq.~\eqref{eq:rec_loss}), which weights every activation direction equally, with the functional error $\bm{e}\hat{\bm{F}}\bm{e}^{\top}$ (Eq.~\eqref{eq:fisher_quad}), which counts only errors that change the policy's action. The functional error alone would leave the null space of $\hat{\bm{F}}$ unconstrained, so we add a small isotropic term. Fisher-weighted featurizers minimize $\mathbb{E}\big[\bm{e}\,(\hat{\bm{F}}/\bar\lambda+\delta\bm{I})\,\bm{e}^{\top}\big]$, where $\bar\lambda$ is the mean eigenvalue of $\hat{\bm{F}}$ and $\delta=0.25$ keeps one fifth of the weight as plain MSE. We implement this by training an unchanged featurizer on whitened activations $\bm{x}\bm{W}_{\!f}$ with $\bm{W}_{\!f}=(\hat{\bm{F}}/\bar\lambda+\delta\bm{I})^{1/2}$, and map decoded features and steering edits back to the residual stream with $\bm{W}_{\!f}^{-1}$.

We train Fisher-weighted featurizers for MolmoAct2 at every odd block up to 33. For a 2,000 DROID episode subset, we compute the gradient of the teacher-forced log-likelihood of the on policy action tokens with respect to the residual stream, yielding roughly 28M activation--gradient pairs per layer, and clip gradient norms to their $[p_{0.5},\,p_{99.5}]$ range.

\end{document}